\documentclass[10pt,twocolumn,letterpaper]{article}

\usepackage{cvpr}
\usepackage{times}
\usepackage{epsfig}
\usepackage{graphicx}
\usepackage{amsmath}
\usepackage{amssymb}

\usepackage{newtxtext}
\newcommand{\comment}[1]{}

\def\bS{\mathbf{S}}

\def\bq{\mathbf{q}}

\def\blambda{\mathbf{\lambda}}

\def\btheta{\boldsymbol{\theta}}

\def\V{\mathcal{V}}

\def\bI{\mathbf{I}}
\def\H{\mathcal{H}}

\def\bS{\mathbf{S}}

\def\bI{{\bf I}}

\usepackage{multirow}
\usepackage{multicol}

\def\red#1{\textcolor{red}{#1}}

\def\bepsilon{\boldsymbol{\epsilon}}

\def\ie{{\em i.e.}}
\def\eg{{\em e.g.}}

\usepackage{fixltx2e}

\usepackage[nopar]{lipsum}

\usepackage[pagebackref=true,breaklinks=true,letterpaper=true,colorlinks,bookmarks=false]{hyperref}

\cvprfinalcopy % *** Uncomment this line for the final submission

\usepackage{authblk}
\newcommand*\samethanks[1][\value{footnote}]{\footnotemark[#1]}

\def\cvprPaperID{1560} % *** Enter the CVPR Paper ID here

\ifcvprfinal\fi
\begin{document}

%%%%%%%%% TITLE
\title{Exact Adversarial Attack to Image Captioning  \\ via Structured Output Learning with Latent Variables}

\author{
Yan Xu\textsuperscript{${\ddag}$${\dag}$}\thanks{indicates equal contributions. ${^\sharp}$indicates corresponding authors. This work was done when Yan Xu was an intern at Tencent AI Lab.},~  Baoyuan Wu\textsuperscript{${\dag}$${\sharp}$}\samethanks, ~ Fumin Shen\textsuperscript{${\ddag}$}, Yanbo Fan\textsuperscript{${\dag}$}, Yong Zhang\textsuperscript{${\dag}$}, Heng Tao Shen\textsuperscript{${\ddag}$}, Wei Liu\textsuperscript{${\dag}$${\sharp}$}
\\
\textsuperscript{${\dag}$}Tencent AI Lab, 
\textsuperscript{${\ddag}$}University of Electronic Science and Technology of China 
\\
{\tt\small \{xuyan5533,wubaoyuan1987,fumin.shen,fanyanbo$0124$,zhangyong$201303$\}@gmail.com, ~~~~~~~~~~~ %\{yanbofan,norriszhang\}@tencent.com, 
shenhengtao@hotmail.com, wl2223@columbia.edu }
}

\maketitle
\thispagestyle{empty}

%%%%%%%%% ABSTRACT
\begin{abstract}
   In this work, we study the robustness of a CNN+RNN based image captioning system being subjected to adversarial noises. 
   We propose to fool an image captioning system to generate some targeted partial captions for an image polluted by adversarial noises, even the targeted captions are totally irrelevant to the image content.
   A partial caption indicates that the words at some locations in this caption are observed, while words at other locations are not restricted. It is the first work to study exact adversarial attacks of targeted partial captions.
   Due to the sequential dependencies among words in a caption, 
   we formulate the generation of adversarial noises for targeted partial captions as a structured output learning problem with latent variables. Both the generalized expectation maximization algorithm and structural SVMs with latent variables are then adopted to optimize the problem. 
   The proposed methods generate very successful attacks to three popular CNN+RNN based image captioning models.
   Furthermore, the proposed attack methods are used to understand the inner mechanism of image captioning systems, 
   providing the guidance to further improve automatic image captioning systems towards human captioning. 
\end{abstract}

%%%%%%%%% BODY TEXT
\vspace{-1em}
\section{Introduction}
It has been shown \cite{szegedy2013intriguing} that deep neural networks (DNNs) \cite{deep-learning-nature-2015} are vulnerable to adversarial images, which are visually similar to benign images. Most of these works focus on convolutional neural networks (CNNs) \cite{cnn-lecun-1995} based tasks (\eg, image classification \cite{alexnet-nips-2012,my-cvpr-2017,my-cvpr-2018,tencent-ml-images-2018}, object detection \cite{RCNN-CVPR-2014}, or object tracking \cite{zhang2012robust,TADT}),
of which the loss functions are factorized to independent (\ie, unstructured) outputs, so that the gradient can be easily computed to generate adversarial noises. 
However, if the output is structured, it may be difficult to derive the gradient of the corresponding structured loss. 
%It may explain the fact that the adversarial attacks to DNNs with structured outputs have been rarely studied. 
%
One popular deep model with structured outputs is the combination of CNNs and recurrent neural networks (RNNs) \cite{RNN-2015}, where the visual features extracted by CNNs are fed into RNNs to generate a sequential output. 
We call this combination as a CNN+RNN architecture in this paper.
One typical task utilizing the CNN+RNN architecture is image captioning \cite{show-tell-cvpr-2015}, which describes the image content using a sentence. 
In this work, we present adversarial attacks to image captioning, as an early attempt of the robustness of DNNs with structured outputs.

\begin{figure}[t]
\centering
\includegraphics[width=1.01\linewidth, height=2.4in]{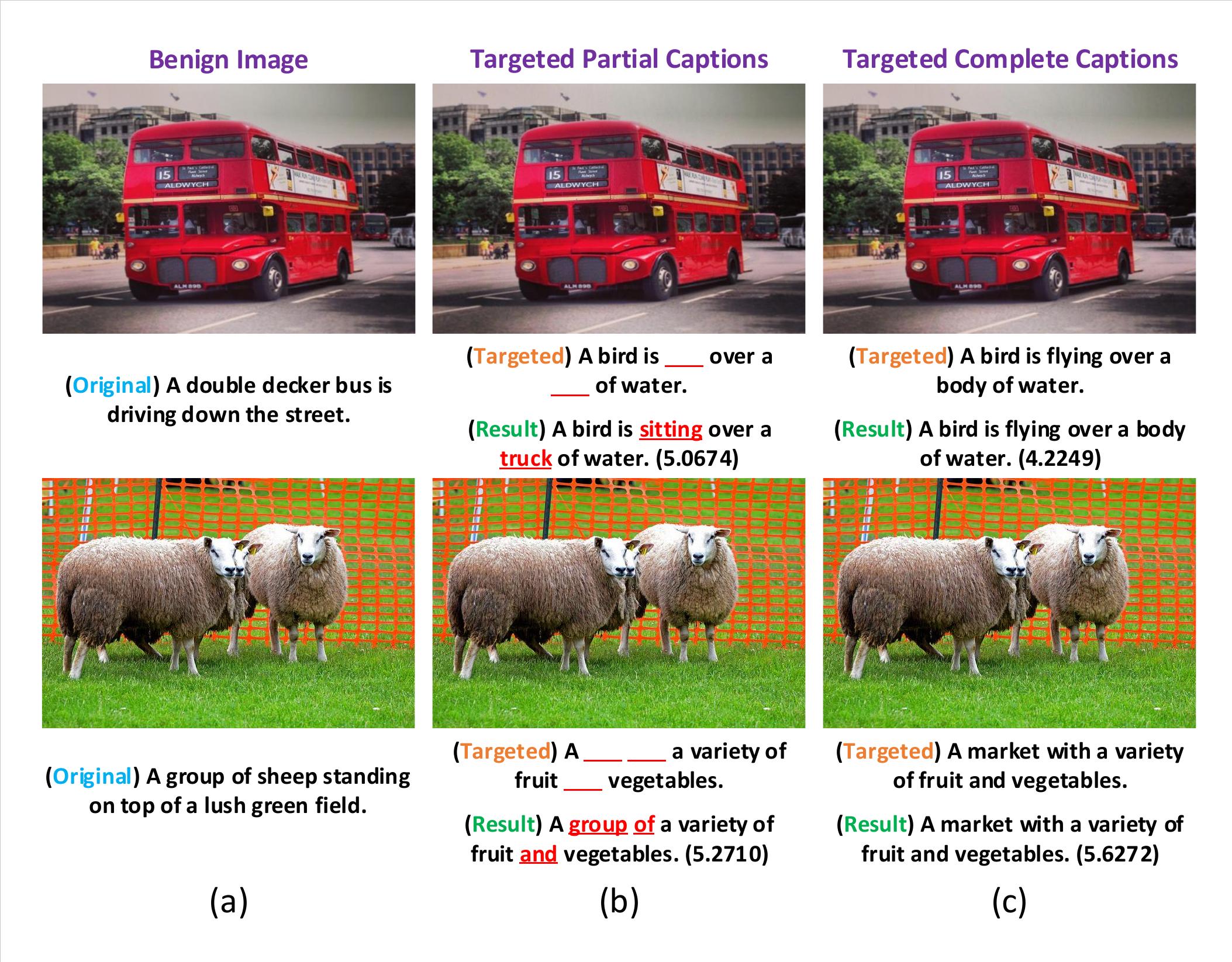}
\vspace{-0.27in}
\caption{ \small
Examples of adversarial attacks to the image captioning model of Show-Attend-and-Tell \cite{show-attend-tell-icml-2015}, using the proposed attack methods dubbed GEM (the top row) and latent SSVMs (the bottom row), respectively. 
In each targeted partial caption (\ie, targeted), the red `$\_$' indicates one latent word.
In each predicted caption (\ie, result), the value at the end denotes the norm of adversarial noises $\| \bepsilon \|_2$. 
All targeted captions are successfully attacked, while adversarial noises are invisible to human perception. 
\comment{
{\bf (a) column}: The benign images and the corresponding captions; 
{\bf (b) column}: The attacks of targeted complete captions, the optimized adversarial images and the predicted captions, with the end of the norm of adversarial noises $\| \bepsilon \|_2$; 
{\bf (c) column}: The attacks of targeted partial captions, the optimized adversarial images and the predicted captions.}
}
\label{fig: motivation example}
\vspace{-0.27in}
\end{figure}

Given a trained CNN+RNN image captioning model and an benign image, 
we want to fool the model to produce a targeted partial caption, which may be totally irrelevant to the image content, through adding adversarial noises to that image. 
This task is called {\it exact adversarial attack of targeted partial captions}, which has never been studied in previous work.
As shown in Fig. \ref{fig: motivation example}(b), a targeted partial caption indicates that the words at some locations are observed, while the words at other locations are not specified, \ie, latent. 
When the words at all locations are observed, it becomes a targeted complete caption (see Fig. \ref{fig: motivation example}(c)).
To this end, the marginal posterior probability of the targeted partial caption should be maximized, while minimizing the norm of adversarial noises. 
It could be formulated as a structured output learning problem with latent variables \cite{structured-output-learning-2007,ssvm-latent-icml-2009}.
Specifically, we present two formulations. % according to two popular models in structured output learning. 
One is maximizing the log marginal likelihood of the targeted partial caption, which can be optimized by the generalized expectation maximization (GEM) algorithm \cite{bishop-2006}. 
The other is maximizing the margin of the log marginal likelihood between the  targeted partial caption and all other possible partial captions at the same locations, which can be optimized by the structural support vector machines with latent variables (latent SSVMs) \cite{ssvm-latent-icml-2009}. 
Note that the proposed formulations %and optimization methods 
are not coupled with any specific CNN+RNN architecture. Thus, we evaluate the proposed methods on three popular image captioning models, including Show-and-Tell \cite{show-tell-cvpr-2015}, Show-Attend-and-Tell \cite{show-attend-tell-icml-2015} and self-critical sequence training (SCST) utilizing reinforcement learning \cite{self-critical-image-caption-cvpr-2017}. 
Experiments on MS-COCO \cite{mscoco-eccv-2014} demonstrate that the proposed methods can generate successful adversarial attacks. % with very high fooling rates. 
As shown in Fig. \ref{fig: motivation example}(b, c), the targeted captions are successfully attacked, while the adversarial noises are invisible to human perception.

It should be emphasized that the value of this work is not just exploring the robustness the image captioning system, but also understanding its inside mechanism. 
%For example, one can design any style of targeted captions, such as passive sentence or attributive clause (see Fig. \ref{fig: passive and attributive}), to check whether it could be produced for a visually natural image (though polluted by adversarial noises).  If such a style cannot be produced, then it means that the corresponding grammar has not been learned by the captioning model.  We also present the attacks of untargeted captions (see Fig. \ref{fig: untargeted}), revealing that the captioning model is likely to produce invalid captions (\ie, violating the grammar of natural languages). 
%These analyses presented in Section \ref{sec: discussions} can be exploited to guide the improvement of the automatic captioning towards to human captioning.
The analyses about untargeted captions and the style of targeted captions
 could reveal the differences between automatic captioning and human captioning, as shown in Section \ref{sec: discussions}.
 %can be exploited to guide the improvement of the automatic captioning model towards to human captioning. 
%
Moreover, the proposed formulation based on structured output learning is independent with any specific task. 
%To the best of our knowledge, it has not been proposed in existing methods.
It provides a new perspective for exact
attacks to deep neural networks with structured outputs, which has not been well studied.

\comment{
%Besides, the proposed method can be used as a probe tool to help the understanding the mechanism of the image captioning system, considering that the CNN+RNN architecture is trained end-to-end. 
Specifically, utilizing the proposed method, we can check that what kind of captions can be produced by the image captioning system, given a visually natural image (though perturbed by adversarial noises). 
For example, as shown in \red{Fig. XXX (a)}, the targeted captions of active voice can be easily produced, while the corresponding captions of passive voice can not be achieved. It demonstrates that the grammar of passive voice has not be learned by the system. 
The other example in \red{Fig. XXX (b)} shows that what kind of phrase combinations has (not) been learned by the system. 
Consequently, a clear picture of the distribution of the output space of the image captioning system could be obtained, which is then used to provide the targeted guidance/supervision to improve the automatic system to be more close to human captioning. 
}

The main contributions of this work are four-fold. 
{\bf (1)} We are the first to study the adversarial attack of targeted partial captions to image captioning systems.
{\bf (2)} We formulate this attack problem as structured output learning with latent variables.
{\bf (3)} Extensive experiments show that state-of-the-art image captioning models can be easily attacked by the proposed methods. 
{\bf (4)} We utilize the attack method to understand the inner mechanism  of image captioning systems. 
%{\bf (4)} We present detailed analysis about the robustness of image captioning systems from multiple perspectives, and provide insight on how to enhance the robustness. 

%-------------------------------------------------------------------------
\vspace{-0.4em}
\section{Related Work}
\label{sec: related work}
\vspace{-0.45em}

Deep neural networks (DNNs) were firstly shown in \cite{szegedy2013intriguing} to be vulnerable to adversarial examples, and many seminal methods have been developed in this literature.
According to the information about the attacked model accessible to the attacker, existing works can be generally partitioned into three categories, including white-box, gray-box, and black-box attacks. 
%According to the attack goal, two general categories of attacks include untargeted and targeted attacks. 
% Besides, another main stream of this literature is studying the defense to adversarial attacks. However, it is out of the scope of this paper. 
We refer the readers to the survey of adversarial examples in \cite{attack-survey-2018} for more details. % of different categories of adversarial attacks and defenses. 
In this section, we categorize existing works according to the outputs of the attacked model, including independent and structured outputs. 

\vspace{0.01em} \noindent
{\bf Adversarial attacks to DNNs with independent outputs.}
Since DNNs (especially CNNs) show very encouraging results on many visual tasks (\eg, image classification \cite{alexnet-nips-2012}, object detection \cite{RCNN-CVPR-2014}, and semantic segmentation \cite{fcn-segmentation-cvpr-2015}), many previous works have also studied the robustness of these DNN-based visual tasks. 
%In the following, we will give a brief review of these works.
For example, image classification is a typical successful visual application of CNNs, and it is also widely studied to verify the newly developed adversarial attack methods, such as box-constrained L-BFGS \cite{szegedy2013intriguing}, fast-gradient-sign method (FSGM) \cite{FSGM-ICLR-2015}, iterative FSGM \cite{I-FSGM-2016}, momentum iterative FSGM\cite{mi-fsgm-cvpr-2018}, Carlini and Wagner attack \cite{CW-attack-2017}, DeepFool \cite{deepfool-cvpr-2016}, \etc. These works demonstrate that image classification based on popular CNN models (\eg, ResNet \cite{resnet-kaiming-cvpr-2016}  or Inception-v3 \cite{inception-v3}) is very vulnerate to adversarial examples. 
%Besides the recognition of general objects, the robustness of CNN-based face recognition is also studied in \cite{attack-to-face-recognition-2016}. 
%
The robustness of other typical visual tasks, \eg, object detection and semantic segmentation, is also studied in \cite{attack-to-detection-arxiv-2017, attack-to-object-detection-arxiv-2017,attack-to-segmentation-detection-iccv-2017} and \cite{attack-to-segmentation-detection-iccv-2017, attack-to-detection-arxiv-2017, attack-segmentation-cvpr-2018}, respectively.
A common trait of above works is that they focus on CNNs and their loss functions are factorized to independent outputs. Consequently, the gradients of the loss function with respect to the input image can be easily computed to generate adversarial noises. 

\vspace{0.01em} \noindent
{\bf Adversarial attacks to DNNs with structured outputs.}
However, the outputs of some deep models are structured. One typical model is the CNN+RNN architecture, of which the output is a temporally dependent sequence. It has been the main-stream model in some visual tasks, such as image captioning \cite{show-tell-cvpr-2015} and visual question answering \cite{VQA-iccv-2015}. 
Due to the dependencies among words in the sequence, it may be difficult to compute the gradient of the attack loss function with respect to the noise. 
An early attempt to attack CNN+RNN based tasks %(including image captioning and visual question answering) 
was proposed in \cite{xu_2018_CVPR}. However, it can only implement attacking of targeted complete sentences, and treat structured outputs as single outputs.
A recent attack to the CNN+RNN based image captioning system is 
%developed in \cite{attack-to-image-caption-acl-2018}, 
called {\it Show-and-Fool} \cite{attack-to-image-caption-acl-2018}. It presents two types of attacks, including targeted captions and targeted keywords. 
Its attack of targeted captions is a special case of our studied attack of targeted partial captions. %, \ie, targeted complete captions.
Its attack of targeted keywords encourages the predicted sentence to include the targeted keywords, but their locations cannot be specified.
In contrast, our attack of targeted partial captions could enforce the targeted keywords to occur at specific locations, which is more restricted.  
Moreover, the formulations and optimization methods of Show-and-Fool are totally different with ours. 
%For example, the loss function of the max margin criteria in Show-and-Fool is factorized to the word at each location (see Section \ref{sec: comparison with ACL 2018} for details), while our structured loss is defined over the whole sequence. 
Its formulations of targeted captions and keywords are different, while the proposed structured output learning with latent variables provides a systematic formulation for both attacks of targeted partial and complete captions.
\comment{
An early attempt to attack a CNN+RNN based image captioning system, called Show-and-Fool, is developed in \cite{attack-to-image-caption-acl-2018}, including two types of attacks, \ie, the targeted caption and targeted keywords. 
Its targeted caption attack is same with the attack of targeted complete captions studied in this work, and a special case of the studied attack of targeted partial captions. 
However, there is a difference between the formulation of Show-and-Fool and our formulation. 
Show-and-Fool adopts the logits (\ie, the inputs of the Softmax function in RNN) in the loss function, while we utilize the log probabilities (\ie, the log of the outputs of the Softmax function in RNN).
The targeted keywords attack in Show-and-Fool encourages the predicted sentence to include the targeted keywords, while the locations of these keywords cannot be specified.
In contrast, our attack of targeted partial captions could enforce the targeted keywords to occur at specific location of the predicted caption, which is more restricted.  
Moreover, a gate function has to be introduced to avoid {\it keyword collision} in the attack of targeted keywords in \cite{attack-to-image-caption-acl-2018}. 
In contrast, our proposed structured output learning with latent variables provides a unified and systematic formulation for both targeted attacks of complete and partial captions. 
}

%-------------------------------------------------------------------------
\vspace{-0.3em}
\section{Structured Outputs of CNN+RNN based Image Captioning Systems}
\label{sec: structured outputs of captioning}
\vspace{-0.3em}

Given a trained CNN+RNN based captioning model with parameters $\btheta$, and an perturbed image $\bI = \bI_0 + \bepsilon \in [0,1]$, the posterior probability of a caption $\bS$ is formulated as
\vspace{-0.25em}
\begin{equation}
\vspace{-0.4em}
P(\bS | \bI_0, \bepsilon; \btheta) = \prod_{t=1}^N 
P(\bS_t | \bS_{<t}, \bI_0, \bepsilon; \btheta),
\label{eq: joint posterior}
\vspace{-0.35em}
\end{equation}
where $\bI_0$ represents the benign image, and $\bepsilon$ denotes the adversarial noise. $\bS = \{ \bS_1, \ldots, \bS_t, \ldots, \bS_N \}$ indicates a sequence of $N$ variables.  
$\bS_t$ indicates the output variable of $t$-step, and its state could be one from the candidate set $V = \{1, 2, \ldots, |\V| \}$, corresponding to the set of candidate words, \ie, $\V$. 
$\bS_{< t} = \{ \bS_1, \ldots, \bS_{t-1} \}$; 
when $t=1$, we define $\bS_{< t} = \emptyset$. 
Note that we do not specify the formulation of $P(\bS_t | \bS_{<t}, \bI_0, \bepsilon; \btheta)$, and it can be specified as any CNN+RNN model (\eg, Show-and-Tell \cite{show-tell-cvpr-2015}).
For clarity, we ignore the notations $\bI_0$ and $\btheta$ hereafter. 

Besides, a partial caption is denoted as $\bS_{\mathcal{O}}$, which means that the variables at the specific places $\mathcal{O}$ are observed, while other variables are unobserved, \ie, latent. 
Specifically, we define $\mathcal{O} \subset \{1, 2, \ldots, N \}$ and $\bS_{\mathcal{O}} = \{ \bS_t | t \in \mathcal{O} \}$, where $\bS_t = s_t$ with $s_t \in V$ being the observed state. All observed states are summarized as an ordered set $S_{\mathcal{O}} = \{ s_t | t \in \mathcal{O} \}$. 
The latent variables are defined as $\bS_{\mathcal{H}} = \{ \bS_t | t \in \mathcal{H} \equiv \{1, 2, \ldots, N \} \setminus \mathcal{O} \} = \bS \setminus  \bS_{\mathcal{O}}$. 
Then, the posterior probability of the partial caption $\bS_{\mathcal{O}}$ is formulated as:
\begin{equation}
\vspace{-0.5em}
P(\bS_{\mathcal{O}} | \bepsilon) = \sum_{\bS_{\mathcal{H}}} P(\bS_{\mathcal{O}}, \bS_{\mathcal{H}} | \bepsilon),
\vspace{-0.5em}
\end{equation}
where $\sum_{\bS_{\mathcal{H}}}$ indicates the summation over all possible configurations of latent variables $\bS_{\mathcal{H}}$. 

%-------------------------------------------------------------------------
\vspace{-0.3em}
\section{Adversarial Attack of Targeted Partial Captions to Image Captioning}
\label{sec: targeted attack}
\vspace{-0.3em}

%\vspace{0.5em}
\noindent
{\bf Learning $\bepsilon$.}
The goal of the targeted partial caption attack is to enforce the predicted caption $\bS$ to be compatible with $\bS_{\mathcal{O}}$, meaning that the predicted words at $\mathcal{O}$ are exactly $S_{\mathcal{O}}$. 
To this end, while minimizing the norm of adversarial noises, either of the following two criterion can be adopted. 
{\bf (1)} The log marginal likelihood $\ln P(\bS_{\mathcal{O}} = S_{\mathcal{O}} | \bepsilon)$ is maximized (see Section \ref{sec: subsec max likelihood by EM}). 
{\bf (2)} The margin of the log marginal likelihood between the targeted caption (\ie, $\ln P(\bS_{\mathcal{O}} = S_{\mathcal{O}} | \bepsilon)$) and all other possible partial captions (\ie, $\ln P(\hat{\bS}_{\mathcal{O}} \neq S_{\mathcal{O}} | \bepsilon)$) is maximized. It is formulated as structural SVMs with latent variables (see Section \ref{sec: subsec structural svm}).

\noindent
{\bf Inference.} Given the optimized $\bepsilon$, the caption of the image perturbed by $\bepsilon$ is inferred as follows:
\begin{flalign}
\vspace{-0.5em}
\bS^*_{\bepsilon} = \arg\max_{\bS} P(\bS| \bI_0 + \bepsilon).
\label{eq: inference of final caption with noise}
\end{flalign}

\vspace{-0.5em}
\subsection{Maximizing Log Marginal Likelihood via Generalized EM Algorithm}
\label{sec: subsec max likelihood by EM}

According to the first criterion, the adversarial noise $\bepsilon$ for the targeted partial caption is derived by the maximization of log marginal likelihood, while minimizing $\| \bepsilon \|_2^2$, as follows: 
%It is formulated as 
\vspace{-1em}
\begin{align}
\vspace{-2em}
  & \arg\max_{\bepsilon} \ln P(\bS_{\mathcal{O}} | \bepsilon)  - \lambda \| \bepsilon \|_2^2
\label{eq: max log likelihood}
\\
 \vspace{-5em}
 \equiv & \arg\max_{\bepsilon} \ln \sum_{\bS_{\mathcal{H}}} P(\bS_{\mathcal{O}}, \bS_{\mathcal{H}} | \bepsilon) - \lambda \| \bepsilon \|_2^2, 
\nonumber
\vspace{-3.5em}
\end{align}
subject to the constraint $\bI_0 + \bepsilon \in [0, 1]$. 
This constraint can be easily satisfied by clipping. For clarity, we ignore it hereafter.
$\lambda$ denotes the trade-off parameter. 
Due to the summation over all possible configurations of $\bS_{\mathcal{H}}$, the above problem is difficult. 
To tackle it, the generalized expectation maximization (GEM) algorithm \cite{bishop-2006} is adopted. The core idea of GEM is introducing the factorized posterior $\bq(\bS_{\mathcal{H}}) = \prod_{t \in \mathcal{H}} \bq(\bS_t)$ to approximate the posterior probability $P(\bS_{\mathcal{H}} |\bS_{\mathcal{O}}, \bepsilon)$. Then, we have the following equation, 
\vspace{-1em}
\begin{flalign}
\vspace{-1em}
\ln P(\bS_{\mathcal{O}} | \bepsilon) &= \mathcal{L} (\bq, \bepsilon) + KL(\bq \parallel P(\bS_{\mathcal{H}} |\bS_{\mathcal{O}}, \bepsilon)), 
\\
% \vspace{-0.23em}
% \end{equation}
% where 
% \vspace{-1em}
%\begin{equation}
 \mathcal{L}(\bq, \bepsilon)  &= \sum_{\bS_{\mathcal{H}}} \bq(\bS_{\mathcal{H}}) \ln \frac{P(\bS_{\mathcal{O}}, \bS_{\mathcal{H}} | \bepsilon)}{\bq(\bS_{\mathcal{H}})},
 \vspace{-1em}
\end{flalign}
\vspace{-1em}
\begin{equation}
 KL\big(\bq(\bS_{\mathcal{H}}) \parallel P(\bS_{\mathcal{H}} |\bS_{\mathcal{O}}, \bepsilon)\big)  = 
\sum_{\bS_{\mathcal{H}}} \bq(\bS_{\mathcal{H}}) \ln \frac{\bq(\bS_{\mathcal{H}})}{P(\bS_{\mathcal{H}} |\bS_{\mathcal{O}}, \bepsilon)}.
\vspace{-0.23em}
\end{equation}
According to the property of the KL divergence that $KL(\bq(\bS_{\mathcal{H}}) \parallel P(\bS_{\mathcal{H}} |\bS_{\mathcal{O}}, \bepsilon)) \geqslant 0$, we obtain that 
%$\mathcal{L}(\bq, \bepsilon)$ is the lower bound of $\ln P(\bS_{\mathcal{O}} | \bepsilon)$, \ie, 
$\mathcal{L}(\bq, \bepsilon) \leqslant \ln P(\bS_{\mathcal{O}} | \bepsilon)$. 
Consequently, the maximization problem (\ref{eq: max log likelihood}) can be optimized through the following two alternative sub-problems, until convergence. 

\vspace{0.5em}
\noindent
\textbf{E step}: Given $\bepsilon$, $\bq(\bS_{\mathcal{H}})$ is updated by minimizing the following equation
\comment{
\begin{flalign}
& KL\big(\bq(\bS_{\mathcal{H}}) \parallel P(\bS_{\mathcal{H}} |\bS_{\mathcal{O}}, \bepsilon)\big)  = 
\sum_{\bS_{\mathcal{H}}} \bq(\bS_{\mathcal{H}}) \ln \bq(\bS_{\mathcal{H}})  
\\
& 
- \sum_{\bS_{\mathcal{H}}} \bq(\bS_{\mathcal{H}}) \big[ \ln P(\bS_{\mathcal{O}}, \bS_{\mathcal{H}} | \bepsilon) - \ln \sum_{\bS_{\mathcal{H}}} P(\bS_{\mathcal{O}}, \bS_{\mathcal{H}} | \bepsilon) \big] =
\nonumber 
\\
&  \sum_{t=1}^N \sum_{k=1}^{|\V|}  \bq(\bS_t^k)
\big[ \ln \bq(\bS_t^k) - 
\sum_{\bS_{< t, \mathcal{H}}} \bq(\bS_{< t, \mathcal{H}}) \ln P(\bS_t^k | \bS_{< t}, \bepsilon)
\big],
\nonumber 
\end{flalign}
}
\vspace{-0.6em}
\begin{equation}
 KL\big(\bq(\bS_{\mathcal{H}}) \parallel P(\bS_{\mathcal{H}} |\bS_{\mathcal{O}}, \bepsilon)\big)  = 
\sum_{\bS_{\mathcal{H}}} \bq(\bS_{\mathcal{H}}) \ln \bq(\bS_{\mathcal{H}})  
\vspace{-0.5em}
\end{equation}
\vspace{-0.3em}
\begin{equation}
- \sum_{\bS_{\mathcal{H}}} \bq(\bS_{\mathcal{H}}) \big[ \ln P(\bS_{\mathcal{O}}, \bS_{\mathcal{H}} | \bepsilon) - \ln \sum_{\bS_{\mathcal{H}}} P(\bS_{\mathcal{O}}, \bS_{\mathcal{H}} | \bepsilon) \big] =
\nonumber 
\vspace{-0.3em}
\end{equation}
\begin{equation}
\sum_{t=1}^N \sum_{k=1}^{|\V|}  \bq(\bS_t^k)
\big[ \ln \bq(\bS_t^k) - 
\sum_{\bS_{< t, \mathcal{H}}} \bq(\bS_{< t, \mathcal{H}}) \ln P(\bS_t^k | \bS_{< t}, \bepsilon)
\big],
\nonumber 
\vspace{-0.5em}
\end{equation}
where the constant $\sum_{\bS_{\mathcal{H}}} \bq(\bS_{\mathcal{H}}) \big[ \ln \sum_{\bS_{\mathcal{H}}} P(\bS_{\mathcal{O}}, \bS_{\mathcal{H}} | \bepsilon) \big]$ in the last formula is ignored. 
$\bq(\bS_t^k) = \bq(\bS_t = k)$ indicates the probability of the variable $\bS_t$ with the state $k$, and $\sum_{k \in V} \bq(\bS_t^k) = 1$. 
$\bS_{< t} = \{ \bS_1, \ldots, \bS_{t-1} \}$ and 
$\bS_{< t, \mathcal{H}} = \bS_{< t} \cap \bS_{\mathcal{H}}$.
When $t=1$ and $t \in \mathcal{H}$, we define $\bS_{< t, \mathcal{H}} = \emptyset$.
Due to the sequential dependency among $\bS$, the probability 
$\bq(\bS_{t})$ can be updated in an ascending order (\ie, from 1 to $N$). 
%If $t \in \mathcal{O}$, then  $\bq(\bS_{t} = s_t) = 1$ and $\bq(\bS_{t} \neq s_t) = 0$. 
Specifically, with fixed $\bq(\bS_{< t, \mathcal{H}})$, 
the update of $\bq(\bS_t^k)$ is derived by setting its gradient to 0, as follows:
\begin{flalign}
\vspace{-1em}
& 1 + \ln \bq(\bS_t^k) - \sum_{\bS_{< t, \mathcal{H}}} \bq(\bS_{< t, \mathcal{H}}) \ln P(\bS_t^k | \bS_{< t, \mathcal{H}}, \bepsilon) = 0
\nonumber
\end{flalign}
\begin{flalign}
& \Rightarrow 
\bq(\bS_t^k) = \exp\bigg( \sum_{\bS_{< t, \mathcal{H}}} \bq(\bS_{< t, \mathcal{H}}) \ln P(\bS_t^k | \bS_{< t, \mathcal{H}}, \bepsilon) - 1 \bigg)
\nonumber
\\
& \Rightarrow 
\bq(\bS_t^k) \leftarrow 
\bq(\bS_t^k) / \big({\textstyle\sum}_k^{|V|} \bq(\bS_t^k)\big)
%\frac{\bq(\bS_t^k)}{ \sum_k^{|V|} \bq(\bS_t^k)}.
\label{eq: update of q(S_t^k) in E step}
\end{flalign}

\vspace{0.2em}
\noindent
\textbf{M step}: Given $\bq(\bS_{\mathcal{H}})$, $\bepsilon$ is updated as follows:
\begin{flalign}
& \arg\max_{\bepsilon} \mathcal{L}(\bq, \bepsilon) - \lambda \| \bepsilon \|_2^2 
\label{eq: M step in EM}
\\
&= \text{const} - \lambda \| \bepsilon \|_2^2   + \sum_{\bS_{\mathcal{H}}} \bq(\bS_{\mathcal{H}}) \ln P(\bS_{\mathcal{O}}, \bS_{\mathcal{H}} | \bepsilon)  = \text{const} 
\nonumber
\\
& + \sum_{t=1}^N \bigg[ \sum_{\bS_{1 \sim t, \mathcal{H}}} \bq(\bS_{1 \sim t, \mathcal{H}}) \ln P(\bS_t | \bS_{<t}, \bepsilon) \bigg] - \lambda \| \bepsilon \|_2^2, 
\nonumber
\end{flalign}
where $\bS_{1 \sim t, \mathcal{H}} = \{ \bS_1, \ldots, \bS_t \} \cap \bS_{\mathcal{H}}$, and 
$\text{const} = -\sum_{\bS_{\mathcal{H}}} \bq(\bS_{\mathcal{H}}) \ln \bq(\bS_{\mathcal{H}})$. 
It can be easily optimized by any gradient based method for training deep neural networks, such as stochastic gradient descent (SGD) \cite{sgd-1985} or adaptive moment estimation (ADAM) \cite{adam-2014}. It will be specified in our experiments. 
However, the number of all possible configurations of $\bS_{1 \sim t, \mathcal{H}}$ is $|\V|^{|\bS_{1 \sim t, \mathcal{H}}|}$. It could be very large even for moderate $|\bS_{1 \sim t, \mathcal{H}}|$. 
Fortunately, since $\bq(\bS_t^k) \in [0,1]$ and $\sum_k^{|\V|} \bq(\bS_t^k) = 1$ for any $t \in \{1, \ldots, N\}$, the values of $\bq(\bS_{1 \sim t, \mathcal{H}})$ for most configurations of $\bS_{1 \sim t, \mathcal{H}}$ are so small that they can be numerically ignored. 
Thus, we only consider the configurations of top-3 probabilities of $\bq(\bS_t)$ for each latent variable $\bS_t$. Consequently, the number of all configurations is reduced to $3^{|\bS_{1 \sim t, \mathcal{H}}|}$, over which the summation becomes tractable. 

\subsection{Structural SVMs with Latent Variables}
\label{sec: subsec structural svm}

According to the second criteria, the adversarial noise $\bepsilon$ is generated by structural SVMs with latent variables \cite{ssvm-latent-icml-2009}, 
\begin{flalign}
\vspace{-0.2em}
\arg\min_{\bepsilon} & ~ \lambda \| \bepsilon \|_2^2 - \max_{\bS_{\mathcal{H}}} \ln P(\bS_{\mathcal{O}}, \bS_{\mathcal{H}} | \bepsilon)
 \label{eq: max margin}
\\
& + 
\max_{\hat{\bS}_{\mathcal{O}}, \hat{\bS}_{\mathcal{H}}} \big[ \ln P( \hat{\bS}_{\mathcal{O}}, \hat{\bS}_{\mathcal{H}} | \bepsilon) +  \bigtriangleup(\bS_{\mathcal{O}}, \hat{\bS}_{\mathcal{O}})  \big],
\nonumber
\\
\bigtriangleup(\bS_{\mathcal{O}}, \hat{\bS}_{\mathcal{O}}) & = 
%\frac{1}{|\mathcal{O}|} 
\sum_{t \in \mathcal{O}} \bigtriangleup(\bS_{t}, \hat{\bS}_{t}), ~
\bigtriangleup(\bS_{t}, \hat{\bS}_{t}) = 
\begin{cases}
\zeta, & \bS_{t} \neq \hat{\bS}_{t} \\
0, & \bS_{t} = \hat{\bS}_{t},
\end{cases}
\vspace{-1.5em}
\label{eq: structured loss}
\vspace{-1.5em}
\end{flalign}
\comment{
\begin{flalign}
\vspace{-0.2em}
\arg\min_{\bepsilon} & ~ \lambda \| \bepsilon \|_2^2 + 
\max_{\hat{\bS}_{\mathcal{O}}, \hat{\bS}_{\mathcal{H}}} \big[ \ln P( \hat{\bS}_{\mathcal{O}}, \hat{\bS}_{\mathcal{H}} | \bepsilon) +  \bigtriangleup(\bS_{\mathcal{O}}, \hat{\bS}_{\mathcal{O}})  \big]
\nonumber 
\\
& - \max_{\bS_{\mathcal{H}}} \ln P(\bS_{\mathcal{O}}, \bS_{\mathcal{H}} | \bepsilon),
\label{eq: max margin}
\vspace{-0.2em}
\end{flalign}
where
\begin{flalign}
\vspace{-0.2em}
\bigtriangleup(\bS_{\mathcal{O}}, \hat{\bS}_{\mathcal{O}}) = 
%\frac{1}{|\mathcal{O}|} 
\sum_{t \in \mathcal{O}} \bigtriangleup(\bS_{t}, \hat{\bS}_{t}), ~
\bigtriangleup(\bS_{t}, \hat{\bS}_{t}) = 
\begin{cases}
\zeta, & \bS_{t} \neq \hat{\bS}_{t} \\
0, & \bS_{t} = \hat{\bS}_{t}
\end{cases}
\label{eq: structured loss}
\vspace{-1em}
\end{flalign}
}
where the scalar $\zeta > 0$ will be specified in experiments. 
This problem can be optimized by the following two alternative sub-problems, until convergence.

\vspace{0.5em}
\noindent
{\bf (1) Latent variable completion} with fixed $\bepsilon$:
\vspace{-0.2em}
\begin{equation}
\bS_{\mathcal{H}}^* = \arg\max_{\bS_{\mathcal{H}}} \ln P(\bS_{\mathcal{O}}, \bS_{\mathcal{H}} | \bepsilon).
\label{eq: latent variable completion}
\vspace{-0.25em}
\end{equation}
It is solved by sequential inference in an ascending order.
%This problem can be easily solved by sequential inference in an ascending order. 

\vspace{0.5em}
\noindent
{\bf (2) Optimizing $\bepsilon$} via Structural SVMs with fixed $\bS_{\mathcal{H}}^*$:
\begin{flalign}
\arg\min_{\bepsilon} & ~ \lambda \| \bepsilon \|_2^2 + 
\max_{\hat{\bS}_{\mathcal{O}}, \hat{\bS}_{\mathcal{H}}} \big[ \ln P( \hat{\bS}_{\mathcal{O}}, \hat{\bS}_{\mathcal{H}} | \bepsilon) +  \bigtriangleup(\bS_{\mathcal{O}}, \hat{\bS}_{\mathcal{O}})  \big]
\nonumber 
\\
& - \ln P(\bS_{\mathcal{O}}, \bS_{\mathcal{H}}^* | \bepsilon).
\end{flalign}
This problem is also optimized by two alternative steps.

\noindent
{\bf (2.1) Loss augmented inference} with fixed $\bepsilon$:
\vspace{-0.25em}
\begin{equation}
\hat{\bS}_{\mathcal{O}}^*, \hat{\bS}_{\mathcal{H}}^* =  \underset{{\hat{\bS}_{\mathcal{O}}, \hat{\bS}_{\mathcal{H}}}}{\arg\max}
\ln P( \hat{\bS}_{\mathcal{O}}, \hat{\bS}_{\mathcal{H}} | \bepsilon) +  \bigtriangleup(\bS_{\mathcal{O}}, \hat{\bS}_{\mathcal{O}}).
\label{eq: loss augmented inference}
\vspace{-0.25em}
\end{equation}
This inference problem is also sequentially solved in an ascending order. 
Specifically, given the inferred configurations $\hat{\bS}_{<t}^*$, the inference over $\hat{\bS}_t$ is solved as follows:
\begin{enumerate}
    \item When $t \in \mathcal{O}$, $\hat{\bS}_t^* = \arg\max_{\hat{\bS}_t} \big[ \ln P(\hat{\bS}_t | \hat{\bS}_{<t}^*, \bepsilon) + \bigtriangleup(\bS_{t}, \hat{\bS}_{t}) \big]$. 
    \item When $t \in \H$, $\hat{\bS}_t^* = \arg\max_{\hat{\bS}_t}  \ln P(\hat{\bS}_t | \hat{\bS}_{<t}^*, \bepsilon) $.
\end{enumerate}

\comment{
\begin{flalign}
\bigtriangleup(\bS_{\mathcal{O}}, \hat{\bS}_{\mathcal{O}}) & = 
%\frac{1}{|\mathcal{O}|} 
 \sum_{t \in \mathcal{O}} \bigtriangleup(\bS_{t}, \hat{\bS}_{t}), 
\\
\bigtriangleup(\bS_{t}, \hat{\bS}_{t}) 
& = 
\begin{cases}
1, & \bS_{t} \neq \hat{\bS}_{t} \\
0, & \bS_{t} = \hat{\bS}_{t}
\end{cases}
\end{flalign}
}

\noindent
{\bf (2.2) Update $\bepsilon$} with fixed $\hat{\bS}_{\mathcal{O}}^*, \hat{\bS}_{\mathcal{H}}^*$:
\begin{flalign}
\hspace{-0.65em} \arg\min_{\bepsilon} & ~ \lambda \| \bepsilon \|_2^2 + 
\ln P( \hat{\bS}_{\mathcal{O}}^*, \hat{\bS}_{\mathcal{H}}^* | \bepsilon) 
- \ln P(\bS_{\mathcal{O}}, \bS_{\mathcal{H}}^* | \bepsilon).
\label{eq: 2.2 step of update epsilon in SSVM}
\end{flalign}
Similar to M step (see Eq. (\ref{eq: M step in EM})) in GEM, as the gradients of all three terms in the above objective function with respect to $\bepsilon$ can be easily computed, any gradient based optimization method for training deep neural networks can be used. It will be specified in our experiments. 

\vspace{0.5em}
% \noindent
% {\bf Remarks on both Sections \ref{sec: subsec max likelihood by EM} and \ref{sec: subsec structural svm}.} In structured output learning with a general dependency graph (\eg, MRFs \cite{pgm-book-koller}), the involved inference process should be run repeatedly, until convergence. 
% However, in the studied CNN+RNN architecture (\ie, $P(\bS | \bI_0, \bepsilon; \btheta)$ in Eq. (\ref{eq: joint posterior})), the output variables are sequentially dependent, and each variable only depends on all its previous output variables. 
% %It means that the inferred state of latter outputs will not influence the former outputs. 
% Consequently, the inference process needs only one pass along the prediction sequence of RNNs, involving E step (see Eq. (\ref{eq: update of q(S_t^k) in E step})) of GEM, the latent variable completion (see Eq. (\ref{eq: latent variable completion})) and the loss augmented inference (see Eq. (\ref{eq: loss augmented inference})) in latent SSVMs. 
% \red{Excluding the forward and backward through CNNs, the complexities of GEM and latent SSVMs are $O(T(|\V| N^2 + 3^N d))$ and $O( T_{outer}(|\V| N_{\mathcal{H}} + T_{inner}(|\V| N + 2Nd)))$, respectively, with $T$ being the iteration number, $|\V|$ being the vocabulary size, $N$ being the caption length, and $d$ being the output dimension of RNNs.}

\noindent
{\bf Remarks on both Sections \ref{sec: subsec max likelihood by EM} and \ref{sec: subsec structural svm}.} Unlike the general structured output learning with a repeated inference process (\eg, MRFs \cite{pgm-book-koller}), the proposed GEM and latent SSVMs are based on CNN+RNN architecture (\ie, $P(\bS | \bI_0, \bepsilon; \btheta)$ in Eq. (\ref{eq: joint posterior})), which requires only one pass along the prediction sequence of RNNs.
Excluding the forward and backward through CNNs, the complexities of GEM and latent SSVMs are $O\big(T(|\V| N^2 + 3^N d)\big)$ and $O\big( T_{outer}(|\V| N_{\mathcal{H}} + T_{inner}(|\V| N + 2Nd))\big)$, respectively, with $T$ being the iteration number, 
%$|\V|$ being the vocabulary size, $N$ being the caption length, 
and $d$ being the output dimension of RNNs.

%-------------------------------------------------------------------------
\vspace{-0.25em}
\section{Experiments}
\label{sec: experiments}
% this section, we do extensive experiments on different setting to evaluate the effectiveness of the proposed two attack methods. We also compare the proposed algorithm with Show-and-Fool \cite{attack-to-image-caption-acl-2018}, which is the only attack method on image caption as we know. 

\subsection{Experimental Setup}
\label{sec: experimental setup}

In this section, we evaluate the attack performance of the proposed two methods on three CNN+RNN based image captioning models, including Show-Attend-and-Tell (SAT) \cite{show-attend-tell-icml-2015}, self-critical sequence training (SCST) \cite{self-critical-image-caption-cvpr-2017}, and Show-and-Tell (ST) \cite{show-tell-cvpr-2015}. 
We also compare with the only related method (to the best of our knowledge), called Show-and-Fool \cite{attack-to-image-caption-acl-2018}, that also attacks the Show-and-Tell model. 

\vspace{0.2em}
\noindent
{\bf Database and targeted captions.} Our experiments are conducted on the benchmark database for image captioning, \ie, Microsoft COCO 2014 (MSCOCO) \cite{mscoco-eccv-2014}. 
We adopt the split of MSCOCO in \cite{split-of-coco-2015}, including $113,287$ training, $5,000$ validation and $5,000$ test images. 
Following the setting of \cite{attack-to-image-caption-acl-2018}, we randomly select $1,000$ from $5,000$ validation images as the attacked images. 
%We train all three attacked models using the training set of MSCOCO.
Using each attacked model (\ie, SAT, SCST, or ST), we predict the captions of the remaining $4,000$ benign  validation images. We randomly choose $5$ different targeted complete captions from these $4,000$ captions for each attacked image. 
Based on each targeted complete caption, we also generate 6 targeted partial captions, including the partial captions with 1 to 3 latent words (all other words are observed), and those with 1 to 3 observed words (all other words are latent), respectively. Latent or observed words are randomly chosen from each targeted caption. 
As the first word in most targeted captions is `{\it a}', we keep it as observed, and skip it when choosing latent or observed words. 
Due to the memory limit of GPUs, observed words are randomly chosen from the second to the $7^{th}$ location in each targeted caption.
The selected $1,000$ images and corresponding $5,000$ targeted complete captions of each attacked model 
will be released along with our codes in early future. 
%are provided as {\bf supplementary materials}. 

\comment{
For fairness, we randomly select 1,000 images from MSCOCO validation set as the targets to attack. Since the caption model is only able to output relevant captions learned from the training set \cite{attack-to-image-caption-acl-2018}, the targeted captions are generated by the other images in MSCOCO validation set. For reducing contingency, we randomly select $5$ different targeted captions for one target to attack.
}

\comment{
{\bf Evaluation metrics} We measure the performance of the effectiveness of the attack methods using successful rate, average l2 norm of noise, precision and recall. 

The successful rate is mainly to measure the consistency between the observed words and the caption after attack, if the all observed words are generated successfully, this attack is successful. the average l2 norm is used to response the level of noise. these two metrics are also used in \cite{attack-to-image-caption-acl-2018}.

In our work, we report two new metrics, precision and recall, we define them as follow,
\begin{flalign}
Precision = \frac{right\;words\;on\;observed\;location}{observed\;words\;of\;targeted\;caption} \\
Recall = \frac{right\;words\;on\;observed\;location}{observed\;words\;of\;predicted\;caption}
\end{flalign}
These two metrics are proposed to evaluate the completeness between targeted caption and predicted caption.
}

\vspace{0.2em}
\noindent
{\bf Evaluation metrics.} 
Given one targeted caption $\bS_{\mathcal{O}}$ for the benign image $\bI_0$, the adversarial 
noise $\bepsilon$ is measured by its $\ell_2$ norm, \ie, $ \| \bepsilon \|_2$; 
the predicted caption $\bS_{\bepsilon}^*$ (see Eq. (\ref{eq: inference of final caption with noise})) for $\bI_0 + \bepsilon$ is evaluated by the following 
three metrics. 
First, the success sign is defined as follows:
\vspace{-0.25em}
\begin{equation}
\text{succ-sign} = \begin{cases}
1, ~~ \text{if} ~\bS_{\bepsilon, \mathcal{O}}^* \equiv \bS_{\mathcal{O}}
\\
0, ~~ \text{if} ~\bS_{\bepsilon, \mathcal{O}}^* \not\equiv \bS_{\mathcal{O}}
\end{cases},
\vspace{-0.25em}
\end{equation}
where $\equiv$ exactly compares two sequences, and $\bS_{\bepsilon, \mathcal{O}}^* \subset \bS_{\bepsilon}^*$ denotes the sub-sequence of $\bS^*_{\bepsilon}$ at observed locations $\mathcal{O}$.
As $\bS_{\bepsilon}^*$ may be too short to include all observed locations, we know that $|\bS_{\bepsilon, \mathcal{O}}^* | \leqslant |\bS_{\mathcal{O}}|$, with $|\cdot|$ calculating the length of sequence. 
However, succ-sign cannot measure how many inconsistent words in $\bS_{\bepsilon}^*$ with $\bS_{\mathcal{O}}$. 
Thus, we also define the following two metrics:
\vspace{-0.2em}
\begin{equation}
\text{Precision} = \frac{ |\bS_{\bepsilon, \mathcal{O}}^* \cap \bS_{\mathcal{O}} |  }{|\bS_{\bepsilon, \mathcal{O}}^*|}, 
~
\text{Recall} = \frac{ |\bS_{\bepsilon, \mathcal{O}}^* \cap \bS_{\mathcal{O}} |  }{|\bS_{\mathcal{O}}|},
\vspace{-0.2em}
\end{equation}
 where the operator $\cap$ between two sequences returns a sub-sequence including the same words at the same locations. 
 %As $|\bS_{\bepsilon, \mathcal{O}}^* | \leqslant |\bS_{\mathcal{O}}|$, we obtain that $\text{Precision} \geqslant \text{Recall}$.
 % $|\cdot|$ calculates the length of the sequence. 
 If succ-sign is $1$, then both Precision and Recall are $1$; if succ-sign is $0$, then Precision and Recall may be larger than $0$. 
 Besides, considering that $|\bS_{\bepsilon, \mathcal{O}}^* | \leqslant |\bS_{\mathcal{O}}|$, we obtain that $\text{Precision} \geqslant \text{Recall} \geqslant \text{succ-sign}$.
We report the average values of above four metrics over all targeted (partial) captions of all images, \ie, $5000$ captions. The average value of succ-sign is called as success rate (SR). The lower average norm $\| \bepsilon \|_2$, while the higher average values of other three metrics, indicate the better attack performance. 

\vspace{0.2em}
\noindent
{\bf Implementation details.} 
The PyTorch implementations of three target models are downloaded from an open-source GitHub project\footnote{\url{https://github.com/ruotianluo/self-critical.PyTorch}}. 
We train these models based on the training set of MSCOCO. 
We adopt the ResNet-101 architecture \cite{resnet-kaiming-cvpr-2016} as the CNN part in SAT and SCST. Besides, to fairly compare with the Show-and-Fool algorithm \cite{attack-to-image-caption-acl-2018}, we adopt the Inception-v3 \cite{inception-v3} architecture as the CNN part in the ST model. 
For the GEM based attack method, the maximum number of iterations between E and M step is set to $50$; for the latent SSVM based attack method, the maximum numbers of both outer and inner iterations are set to $10$. 
In the M step (see Eq. (\ref{eq: M step in EM})) of GEM, and the $(2.2)$ step (see Eq. (\ref{eq: 2.2 step of update epsilon in SSVM})) of latent SSVMs,  we adopt the ADAM optimization algorithm \cite{adam-2014} to update the noise $\bepsilon$, with the learning rate $0.001$, while all other hyper-parameters are set to the default values in the master branch of PyTorch\footnote{\url{https://github.com/PyTorch/PyTorch/blob/master/torch/optim/adam.py}}. 
If without specific illustrations, the trade-off parameters $\blambda$ in both Eq. (\ref{eq: max log likelihood}) and (\ref{eq: max margin}) are set to $0.1$ in experiments. 
The scalar $\zeta$ of the structured loss (see Eq. (\ref{eq: structured loss})) in latent SSVMs is set to $1$.

\comment{
{\bf Setup} We evaluate the proposed two algorithms on three image caption models, including Show-and-Tell \cite{show-tell-cvpr-2015}, Show-Attend-and-Tell \cite{show-attend-tell-icml-2015} and SCST \cite{self-critical-image-caption-cvpr-2017}. We use the pre-trained PyTorch implementation\footnote{\url{https://github.com/ruotianluo/self-critical.PyTorch}.}. We use Inception-v3 as the CNN for visual feature extraction of Show-and-Tell model, and resnet-101 as the visual feature extraction of Show-Attend-and-Tell and SCST.
For two attack algorithms, we both use ADAM to minimize the loss functions, and we set the learning rate to 0.001 and $\lambda$ to 0.01 without additional illustration. We mainly do experiments with targeted caption (0 latent words), $1 \sim 3$ latent words and $1 \sim 3$ observed words. In one hand, for all generated captions, we observe that the first word is always `a', so we select latent (observed) words from second output, in the other hand, due to the limitations of the GPU memory, we randomly select observed words $1 \sim 3$ from first six words of a caption.
}

\begin{table}[t] %\scriptsize
%\vspace{-0.1in}
\begin{center}
\scalebox{0.65}{
\begin{tabular}{p{.046\textwidth}  p{.055\textwidth} | p{.058\textwidth} | p{.058\textwidth} p{.058\textwidth} p{.058\textwidth} | p{.058\textwidth} p{.058\textwidth} p{.058\textwidth} }
\hline
method & metric & 0 latent & 1 latent & 2 latent & 3 latent & 1 obser & 2 obser & 3 obser
\\
\hline
 \multirow{4}{*}{GEM} &  $\| \bepsilon \|_2$ $\downarrow$ & 4.2767  & 4.4976 & 4.6942 & 4.858 & 3.0304 & 3.5611 & 3.6583
 \\
 & SR ~~~$\uparrow$ & 0.9926 & 0.9154 & 0.759 & 0.5604 & 0.8908 & 0.862 & 0.892
 \\
 & Prec ~$\uparrow$ & 0.9953 &   0.9575 & 0.9092 & 0.856 &  0.8908 & 0.8897 & 0.9236
 \\
 & Rec ~~$\uparrow$ & 0.9953 &   0.9528 & 0.8855 & 0.8 & 0.8908 & 0.8876 & 0.9234
 \\
 \hline
  &  $\| \bepsilon \|_2$ $\downarrow$ & 5.1678 &   5.4558 & 5.7074 & 5.8706 & 5.2509 & 5.6838 & 5.8681
 \\
 Latent & SR ~~~$\uparrow$ & 0.9806 &  0.9126 & 0.8466 & 0.7526 & 0.85 &   0.731 &  0.708
 \\
 SSVMs & Prec ~$\uparrow$ & 0.9892 &   0.955 &  0.9197 & 0.8868 & 0.85 &   0.8092 & 0.8096 
 \\
  & Rec ~~$\uparrow$  & 0.9889 & 0.9524 & 0.9151 & 0.8792 & 0.85 &   0.7896 & 0.7917 
 \\
 \hline
\end{tabular}
}
\end{center}
\vspace{-0.1in}
\caption{ Results of adversarial attack to the Show-Attend-and-Tell model. 
`1 obser' indicates the targeted partial caption of one observed word. 
`Prec' indicates Precision, while `Rec' means Recall. 
$\downarrow$ means that the lower value of that metric is the better attack performance, while 
$\uparrow$ means that the higher value of that metric is the better attack performance.
}
\label{table: results of show attend and tell}
\vspace{-0.17in}
\end{table}

\begin{figure*}
\centering
\includegraphics[width=1\linewidth,height=2.1in]{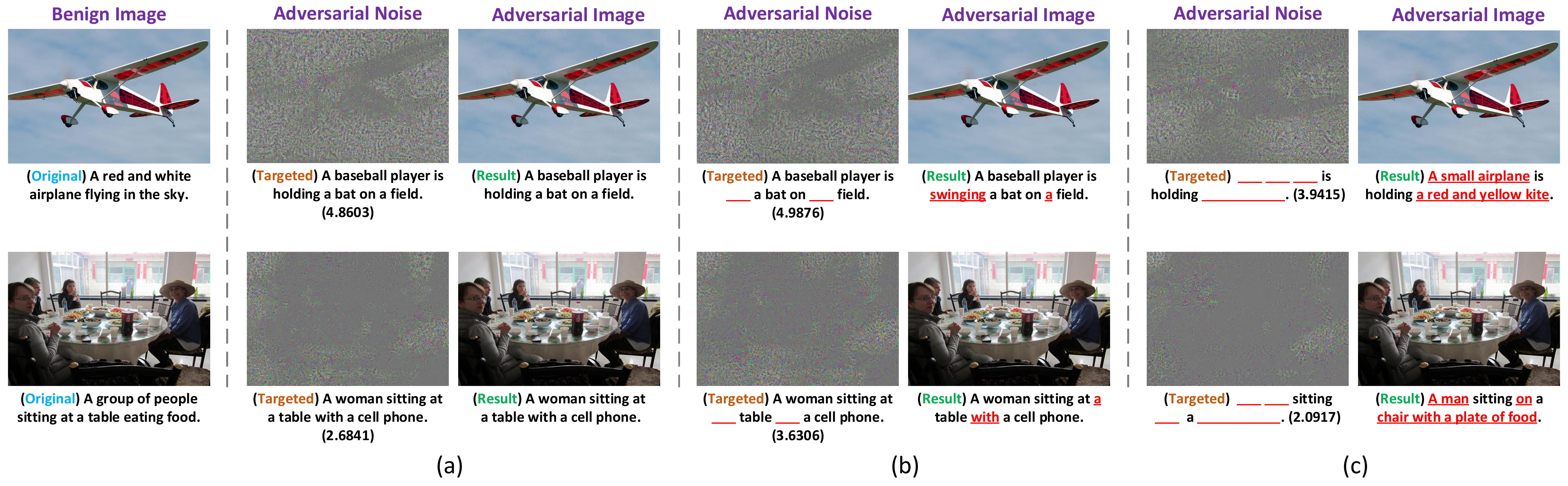}
\vspace{-1.8em}
\caption{Some qualitative examples of adversarial attacks to the Show-Attend-and-Tell \cite{show-attend-tell-icml-2015} model, using the proposed GEM method. Attacks of {\bf (a)} targeted complete captions; {\bf (b)} targeted partial captions with two latent words; {\bf (c)} targeted partial captions with two observed words.
All targeted partial/complete captions are successfully attacked, while the adversarial noises are invisible to human perception.}
\label{fig: visual examples of GEM}
\vspace{-0.13in}
\end{figure*}

\comment{
\subsection{Attack Results of the Show-Attend-and-Tell Model \cite{show-attend-tell-icml-2015}}
\label{Results with Show-Attend-and-Tell model}
}

\subsection{Attack Results of Three State-of-the-Art Image Captioning Models}
\label{sec: results with three models}

%\vspace{0.1em}
\noindent
{\bf Attack results of the Show-Attend-and-Tell model \cite{show-attend-tell-icml-2015}}
%The attack results to the Show-Attend-and-Tell model 
are presented in Table \ref{table: results of show attend and tell}. 
{\bf (1)} In terms of the attacks of targeted complete captions (\ie, `0 latent' in the third column of Table \ref{table: results of show attend and tell}), the SR of GEM is up to $0.9926$, while means that only $37$ targeted captions out of $5,000$ targeted captions are not successfully predicted after generating adversarial noises. And, the corresponding Precision and Recall of GEM are up to $0.9953$. It means that even in failed attacks, many words are also successfully predicted. 
%Later we will present some qualitative results in Fig. \red{XXX}. 
The average noise norm $\| \bepsilon \|_2$ of GEM is $4.2767$. As shown in Fig. \ref{fig: visual examples of GEM}, such small noises are invisible to human perception. %
In contrast, the results of latent SSVMs are slightly worse than those of GEM. 
{\bf (2)} In terms of the attacks of targeted partial captions with 1 to 3 latent words, along the increase of the number of latent words, the results of both GEM and latent SSVMs get worse, with decreasing (SR, Precision, Recall) and increasing $\| \bepsilon \|_2$. 
The reason is that more latent words bring in more uncertainties on predictions of these latent locations. Then, the observed words after latent locations will be influenced by these uncertainties. 
{\bf (3)} In terms of the attacks of targeted partial captions with 1 to 3 observed words, there is not a clear relationship between the attack performance and the number of observed words. 
The reason is that there is a trade-off between satisfying observed words and the uncertainty from the latent words. 
{\bf (4)} In comparison of GEM and latent SSVMs, the average norm $\| \bepsilon \|_2$ of adversarial noises produced by GEM is always lower than that produced by latent SSVMs at all cases. The attack performance (evaluated by SR, Precision and Recall) of GEM is also better than that of latent SSVMs at most cases, excluding two cases of 2 and 3 latent words. However, based on these results, we cannot simply conclude which method is better for adversarial attacks to image captioning. Because these two methods are influenced by the trade-off parameter $\blambda$, and latent SSVMs is also affected by the parameter $\zeta$ defined in Eq. (\ref{eq: structured loss}).

\begin{table}[t] %\scriptsize
%\vspace{-0.1in}
\begin{center}
\scalebox{0.75}{
\begin{tabular}{p{.046\textwidth}  p{.055\textwidth} | p{.055\textwidth}  p{.055\textwidth} p{.055\textwidth} p{.055\textwidth}  p{.058\textwidth} p{.058\textwidth}  }
\hline
\multirow{2}{*}{method} & \multirow{2}{*}{metric} & \multicolumn{6}{c}{$\blambda$}  
\\
  &  &  0.001 & 0.01 & 0.1 & 1 & 10 & 100
\\
\hline
 \multirow{4}{*}{GEM} &  $\| \bepsilon \|_2$ $\downarrow$ & 8.6353 & 7.67 &   4.2767 & 1.6862 & 0.7513 & 0.2701
 \\
 & SR ~~~$\uparrow$ & 0.9956 &   0.9952 & 0.9926 & 0.9402 & 0.4126 & 0.0118
 \\
 & Prec ~$\uparrow$ & 0.9973 &   0.9969 & 0.9953 & 0.9595 & 0.5832 & 0.2128
 \\
 & Rec ~~$\uparrow$ & 0.9972 &   0.9969 & 0.9953 & 0.9589 & 0.5754 & 0.2011 
 \\
 \hline
  &  $\| \bepsilon \|_2$ $\downarrow$ & 9.2682 &   8.2134 & 5.1678 & 2.5074 & 1.023 &  0.2939
 \\
 Latent & SR ~~~$\uparrow$ & 0.985 &   0.9818 & 0.9806 & 0.9252 & 0.4144 & 0.012
 \\
 SSVMs & Prec ~$\uparrow$ & 0.9919 &   0.99 &   0.9892 & 0.9588 & 0.6172 & 0.227
 \\
      & Rec ~~$\uparrow$  & 0.9917 & 0.9897 &   0.9889 & 0.9574 & 0.6092 & 0.2161
 \\
 \hline
\end{tabular}
}
\end{center}
\vspace{-0.1in}
\caption{ \small Attack results of targeted complete captions to the Show-Attend-and-Tell model, with different trade-off parameters $\blambda$ (see Eqs. (\ref{eq: max log likelihood}) and (\ref{eq: max margin})). }
\label{table: results of show attend and tell different lambda}
\vspace{-0.25in}
\end{table}

In the above analysis, the trade-off parameters $\blambda$ in both Eqs. (\ref{eq: max log likelihood}) and (\ref{eq: max margin}) are fixed at $0.1$. 
In the following, we explore the influence of $\blambda$ to the attack performance. 
When $\blambda$ becomes larger, the norm of adversarial noises is expected to be smaller, while the loss gets larger, leading to weaker attack performance. 
This point is fully verified by the results in Table \ref{table: results of show attend and tell different lambda}. 
When $\blambda=0.001$, the SR value of GEM is up to $0.9956$, and $\| \bepsilon \|_2$ is $8.6353$; when $\blambda=100$, the SR value of GEM is up to $0.0118$, and $\| \bepsilon \|_2$ is $0.2701$. 
With the same $\blambda$, GEM performs slightly better than latent SSVMs in most cases, with lower $\| \bepsilon \|_2$ and higher SR, Precision, and Recall. 
However, the performance of latent SSVMs may be also influenced by $\zeta$ (see Eq. (\ref{eq: structured loss})). Due to the space limit, it will be studied in the {\bf supplementary material}.

\comment{
Different settings:
\begin{enumerate}
    \item different CNN+RNN models, Show-and-Tell, Show-Attend-and-Tell, SCST
    \item the number of latent variables
    \item given a fixed number of latent variables, different noises for nouns, adjective, predicate, adverbial
    \item the trade-off parameter $\lambda$
\end{enumerate}

Different presentations:
\begin{enumerate}
    \item successful rate 
    \item norm of $\epsilon$ 
    \item evaluation score between failed top-k captions and targeted captions
    \item successful rate of transfer noises, between different CNN+RNN models
\end{enumerate}
}

\comment{
\subsection{Attack Results of the SCST Model \cite{self-critical-image-caption-cvpr-2017}}
\label{Results with SCST}
}

\vspace{0.2em}
\noindent
{\bf Attack results of the SCST model \cite{self-critical-image-caption-cvpr-2017}}
%The results of adversarial attacks to the SCST model 
are shown in Table \ref{table: results of SCST}. 
The phenomenon behind these results is similar with that behind the results of the Show-Attend-and-Tell model. The reason is that the model structures of Show-Attend-and-Tell and SCST are similar that the visual features extracted by CNNs are fed into RNNs at each step.

\begin{table}[t] %\scriptsize
%\vspace{-0.1in}
\begin{center}
\scalebox{0.65}{
\begin{tabular}{p{.046\textwidth}  p{.055\textwidth} | p{.058\textwidth} | p{.058\textwidth} p{.058\textwidth} p{.058\textwidth} | p{.058\textwidth} p{.058\textwidth} p{.058\textwidth} }
\hline
method & metric & 0 latent & 1 latent & 2 latent & 3 latent & 1 obser & 2 obser & 3 obser
\\
\hline
 \multirow{4}{*}{GEM} &  $\| \bepsilon \|_2$ $\downarrow$ & 5.1978 & 5.5643 & 5.8561 & 6.1171 & 4.3749 & 4.8465 & 4.8419
 \\
 & SR ~~~$\uparrow$ & 0.992 & 0.9168 & 0.7438 & 0.5178 & 0.6344 & 0.6372 & 0.7838
 \\
 & Prec ~$\uparrow$ & 0.9956 &   0.9549 & 0.8847 & 0.7788 & 0.6344 & 0.7328 & 0.8543
 \\
 & Rec ~~$\uparrow$ & 0.9956 &   0.9528 & 0.872 &  0.7503 & 0.6344 & 0.7319 & 0.8543 
 \\
 \hline
  &  $\| \bepsilon \|_2$ $\downarrow$ & 4.7005 &   5.0926 & 5.5109 & 5.8674 & 5.989 &  5.7939 & 5.4646
 \\
 Latent & SR ~~~$\uparrow$ & 0.9804 &  0.916 &  0.8576 & 0.7598 & 0.569 &  0.7066 & 0.8294
 \\
 SSVMs & Prec ~$\uparrow$ & 0.9926 &   0.9684 & 0.934 &  0.8835 & 0.6538 & 0.7835 & 0.8815 
 \\
  & Rec ~~$\uparrow$  & 0.9924 & 0.967 &  0.9306 & 0.8784 & 0.6502 & 0.7809 & 0.8801 
 \\
 \hline
\end{tabular}
}
\end{center}
\vspace{-0.1in}
\caption{ Results of adversarial attacks to the SCST model. }
\label{table: results of SCST}
\vspace{-0.22in}
\end{table}

\comment{
\subsection{Attack Results of the Show-and-Tell Model \cite{show-tell-cvpr-2015}}
\label{Results with show tell}
}

\begin{figure}
\centering
\hspace{-0.1in}
\includegraphics[width=1.01\linewidth, height=2.1in]{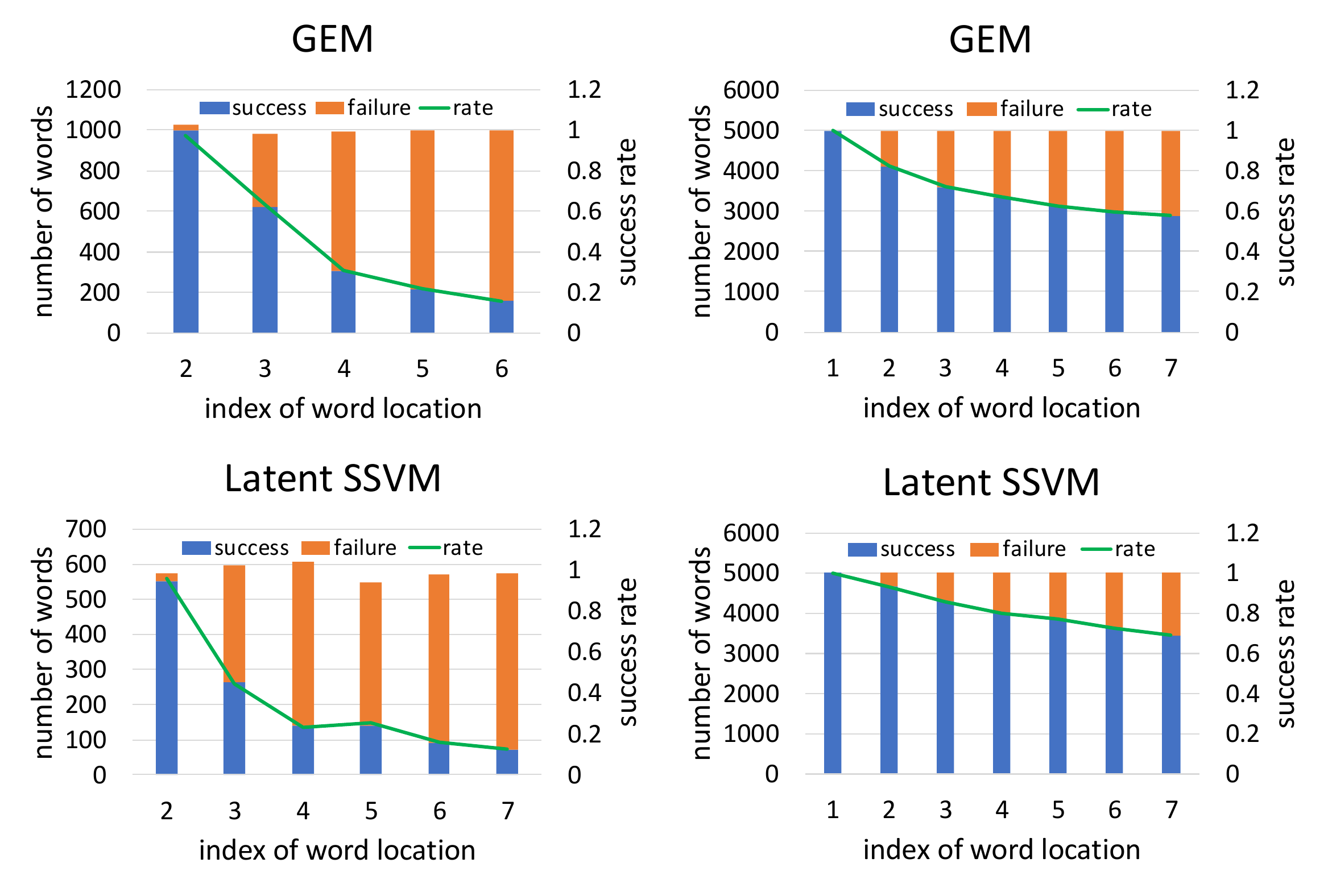}
\vspace{-0.08in}
\caption{ Statistics of success rates of observed words at different locations, on attacking the Show-and-Tell model. {\bf (Left)}: the attack of targeted partial captions with one observed word; {\bf (Right)}: the attack of targeted complete captions. }
\label{fig: SR of location}
\vspace{-0.16in}
\end{figure}

\vspace{0.2em}
\noindent
{\bf Attack results of the Show-and-Tell model \cite{show-tell-cvpr-2015}}
%The results of adversarial attacks to the Show-and-Tell model 
are reported in Table \ref{table: results of show tell}. 
It is found that the attack performance of Show-and-Tell is much worse than that of Show-Attend-and-Tell (see Table \ref{table: results of show attend and tell}) and SCST (see Table \ref{table: results of SCST}). 
The main reason is that the model structure of Show-and-Tell is significantly different with the structures of the other two models. 
Specifically, the visual features extracted by CNN is only fed into the starting step of RNNs, while they are fed into RNNs at every step in Show-Attend-and-Tell and SCST. 
Consequently, the gradients of observed words in targeted partial captions can be directly back-propagated to the input image in Show-Attend-and-Tell and SCST.
In contrast, the gradients of both observed words and latent words are firstly multiplied, then are back-propagated to the input image. Obviously, the influence of observed words becomes much weaker. 
Thus, it is expected that the observed word closer to the end of one caption is more difficult to be successfully attacked. 
To verify this point, we summarize the success rate of observed words at each location. As the lengths of targeted captions vary significantly, we only summarize the words at the first 7 locations. 
As shown in Fig. \ref{fig: SR of location}, in both targeted partial captions with one observed word and targeted complete captions, as well as using both GEM and latent SSVMs, the SR value decreases along the increasing of locations. 

Due to the space limit, we will present: 
(1) attack results of the Show-and-Tell model with the ResNet-101 architecture as the CNN part; 
(2) results of transfer attacks among three captioning models; (3) more qualitative results (like Fig. \ref{fig: visual examples of GEM}) of attacks of targeted partial captions on above three image captioning models, in {\bf supplementary materials}. 
We also report the average runtime of attacking one image in the case of targeted complete captions, using the proposed two methods. 
On the four attacked models, including Show-Attend-and-Tell, SCST, Show-and-Tell with Inception-v3 and Show-and-Tell with ResNet-101, the respective average runtime (seconds) of the GEM method is 95, 81, 36 and 61, and of latent SSVMs is 28, 25, 15 and 24. 
As GEM requires more back-propagation (see the summation term $\sum_{t=1}^N \sum_{\bS_{1 \sim t, \mathcal{H}}}$ in Eq. (\ref{eq: M step in EM})) than latent SSVMs, its runtime is larger.

\begin{table}[t] %\scriptsize
%\vspace{-0.1in}
\begin{center}
\scalebox{0.65}{
\begin{tabular}{p{.046\textwidth}  p{.055\textwidth} | p{.058\textwidth} | p{.058\textwidth} p{.058\textwidth} p{.058\textwidth} | p{.058\textwidth} p{.058\textwidth} p{.058\textwidth} }
\hline
method & metric & 0 latent & 1 latent & 2 latent & 3 latent & 1 obser & 2 obser & 3 obser
\\
\hline
 \multirow{4}{*}{GEM} &  $\| \bepsilon \|_2$ $\downarrow$ & 4.5959 & 3.4488 & 3.3999 & 3.3783 & 2.2588 & 2.5779 & 2.7472
 \\
 & SR ~~~$\uparrow$ & 0.4404 &   0.5034 & 0.4094 & 0.3408 & 0.4606 & 0.4248 & 0.4962
 \\
 & Prec ~$\uparrow$ & 0.6758 &   0.7475 & 0.691 &  0.6455 & 0.4606 & 0.5468 & 0.6403 
 \\
 & Rec ~~$\uparrow$ & 0.6635 &   0.7344 & 0.6763 & 0.626 &  0.4606 & 0.5468 & 0.6403
 \\
 \hline
  &  $\| \bepsilon \|_2$ $\downarrow$ & 1.7635 &   4.5913 & 4.6584 & 4.7369 & 4.5513 & 4.8617 & 4.933
 \\
 Latent & SR ~~~$\uparrow$ & 0.4924 &  0.5808 & 0.4634 & 0.3978 & 0.287 &  0.2118 & 0.227
 \\
 SSVMs & Prec ~$\uparrow$ & 0.7438 &   0.7982 & 0.7257 & 0.6697 & 0.287 &  0.3609 & 0.4065
 \\ 
      & Rec ~~$\uparrow$  & 0.7318 &   0.7862 & 0.7122 & 0.6545 & 0.287 &  0.3459 & 0.3898
 \\
 \hline
\end{tabular}
}
\end{center}
\vspace{-0.1in}
\caption{ Results of adversarial attacks to the Show-and-Tell model. }
\label{table: results of show tell}
\vspace{-0.135in}
\end{table}

\vspace{-0.4em}
\subsection{Comparison with Show-and-Fool \cite{attack-to-image-caption-acl-2018}}
\label{sec: comparison with ACL 2018}
\vspace{-0.4em}

In this section, we compare with the only related work called Show-and-Fool \cite{attack-to-image-caption-acl-2018}. Its attack of targeted captions is a special case of our studied attack of targeted partial captions, \ie, targeted complete captions. %, which is also a special case of our studied attack of targeted partial captions. 
The derivations of two methods proposed in Show-and-Fool also start from the joint probability of a caption given a pre-trained CNN+RNN model, which is same with our derivations. 
However, the derived objective functions of Show-and-Fool are totally different with our objective functions. Specifically, 
\begin{itemize}
\vspace{-0.5em}
    \item {\it Maximizing logits in Show-and-Fool (see Eq. (6) in \cite{attack-to-image-caption-acl-2018})} vs. {\it our maximizing log likelihood (see Eq. (\ref{eq: max log likelihood}))}. Show-and-Fool directly removes the normalization term of the Softmax function of RNNs, as they thought this term is a constant with respect to the input adversarial noises. Actually, this normalization term depends on adversarial noises. Thus, maximizing logits and maxiziming log likelihood are different. 
    \vspace{-0.5em} 
    \item {\it Max margin of logits in Show-and-Fool (see Eq. (7) in \cite{attack-to-image-caption-acl-2018})} vs. {\it our max margin of log likelihood (see Eq. (\ref{eq: max margin}))}. Show-and-Fool maximizes the logit margin between each observed word and all other possible words at the same location. When inferring the word from all other possible words at one location, the corresponding logit exploits the observed word at its previous location as the condition, rather than the inferred word of its previous location. This objective function is standard SVMs factorized at each location, while our objective function is structural SVMs of the whole caption. 
    \vspace{-0.5em}
\end{itemize}

In the following, we present some experimental comparisons between Show-and-Fool and our methods, on the attack of targeted complete captions. 
Show-and-Fool is implemented using Tensorflow\footnote{ \url{https://github.com/huanzhang12/ImageCaptioningAttack}}, and attacks the Show-and-Tell model \cite{show-tell-cvpr-2015}, of which the CNN part is the Inception-v3 model \cite{inception-v3}. To fairly compare with Show-and-Fool, we re-implement our methods using Tensorflow, based on the implementation codes of Show-and-Fool, and attack the same checkpoint of the Show-and-Tell model. 
Besides, Show-and-Fool adopts the arctanh function to transform $\bI_0$ and $\bI_0 + \bepsilon$ to $y=\text{arctanh}(\bI_0), w=\text{arctanh}(\bI_0 + \bepsilon)$, to satisfy the requirement that $\bI_0, \bI_0 + \bepsilon \in [-1,1]$. 
In this section, our method also adopt this setting. 
However, when computing the norm $\| \bepsilon \|_2$ for evaluation, we still transform $\bI_0$ and $\bepsilon$ into the range $\bI_0, \bI_0 + \bepsilon \in [0,1]$. 
%In experiments, we present results of these two settings. 
%
The trade-off parameter $\blambda$ is set to $1$ for both Show-and-Fool and our methods. 
The slack constant $\zeta$ (see Eq. (\ref{eq: structured loss})) in max margin of Show-and-Fool is set to $10,000$ (the default value in the provided code), while $1$ in our SSVM method. 
The experiments are also conducted on the selected 1000 images and 5000 targeted captions (see Section \ref{sec: experimental setup}). The results are shown in Table \ref{table: results of comparison with ACL}. 
Our GEM method shows the best attack performance. 
However, due to the differences on objective functions and optimization methods, these results with similar hyper-parameters may not give a clear conclusion that which method is better. 
But, we still obtain two observations: {\bf (1)} Using logits or log probabilities in the loss term can affect the attack performance;
%{\bf (2)} The input settings of $\bI_0$ and $\bepsilon$ could affect the attack performance; 
{\bf (2)} The comparison between Table \ref{table: results of show tell} and Table \ref{table: results of comparison with ACL} tells that different checkpoints of the same attacked model (\ie, Show-and-tell) will influence the attack performance. 

Show-and-Fool \cite{attack-to-image-caption-acl-2018} also presented the attack of targeted keywords, requiring that the targeted keywords should occur in the predicted caption, but their locations cannot be determined. 
In contrast, our attack of targeted partial captions can enforce the targeted words  to occur at specific locations. 
Besides, the formulations for attacks of targeted captions and targeted keywords are different in Show-and-Fool, while the proposed structured output learning with latent variables provides a systematic formulation of both targeted attacks of complete and partial captions.

\begin{table}[t] %\scriptsize
%\vspace{-0.1in}
\begin{center}
\scalebox{0.69}{
\begin{tabular}{ p{.055\textwidth} | p{.12\textwidth}  p{.17\textwidth} | p{.1\textwidth} p{.12\textwidth}  }
\hline
 \multirow{2}{*}{metric} & \multicolumn{2}{c|}{Show-and-Fool \cite{attack-to-image-caption-acl-2018}}  & \multicolumn{2}{c}{Our methods}
\\
 &  max logits & max margin of logits & GEM & latent SSVMs
 \\
\hline
$\| \bepsilon \|_2$ $\downarrow$ & 1.5202 & 1.7423 & 2.3494 & 4.7854
 \\
 SR ~~~$\uparrow$ & 0.5226 & 0.6586 & 0.7134 & 0.4996
 \\
 Prec ~$\uparrow$ & 0.7239 & 0.8009 & 0.8933 & 0.7335
 \\
 Rec ~~$\uparrow$ & 0.7135 & 0.7926 & 0.886 & 0.7215
 \\
 \hline
\end{tabular}
}
\end{center}
\vspace{-0.05in}
\caption{ \hspace{-0.1in} Comparisons between Show-and-Fool \cite{attack-to-image-caption-acl-2018} and our methods. %See the context in Section \ref{sec: comparison with ACL 2018} for details. 
}
\label{table: results of comparison with ACL}
\vspace{-0.1in}
\end{table}

\vspace{-0em}
\section{Extended Discussions}
\label{sec: discussions} 
\vspace{-0em}

\begin{figure}
\centering
\includegraphics[width=1.02\linewidth,height=1.65in]{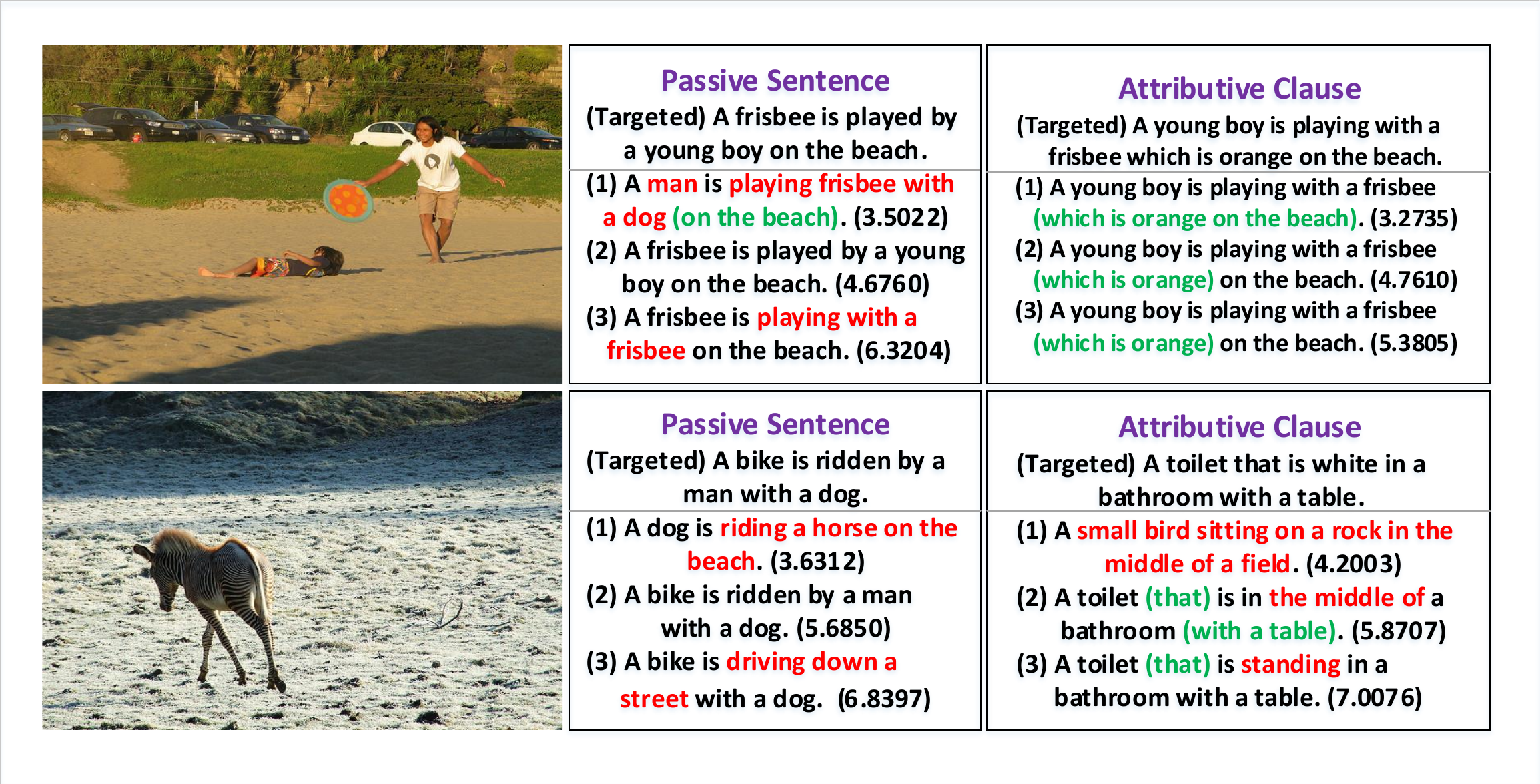}
\vspace{-1.3em}
\caption{Two examples of attacks different styles of targeted captions using GEM. (1), (2) and (3) represent the results of Show-And-Tell, Show-Attend-And-Tell and SCST model, respectively. 
%Note that the targeted captions of the first row match the image content, while those of the second row are irrelevant to the image content. 
%The final value at the end of each predicted caption indicates the average noise norm $\| \bepsilon \|_2$. 
}
\label{fig: passive and attributive}
\vspace{-0.1in}
\end{figure}

\noindent
{\bf What style of targeted captions can be successfully attacked?} 
As demonstrated in Section \ref{sec: experimental setup}, the targeted captions are selected from the captions of 4000 benign validation images. It is found that most of these captions are active sentences, due to that most training captions are active. 
Can the image captioning system produce other styles of captions through adversarial attacking? 
To answer it, we run a simple test of using passive sentences and attributive clauses as targeted captions. As shown in Fig. \ref{fig: passive and attributive} (left), only the Show-Attend-and-Tell model can produce passive sentences, while other two models still keep active. Fig. \ref{fig: passive and attributive} (right) shows that all three models fail to produce attributive clauses. 
It demonstrates that current image captioning systems are not flexible enough to produce different styles of captions like humans.

%\noindent
%{\bf The influence of word type.}

\begin{figure}
\centering
\includegraphics[width=1\linewidth,height=1.6in]{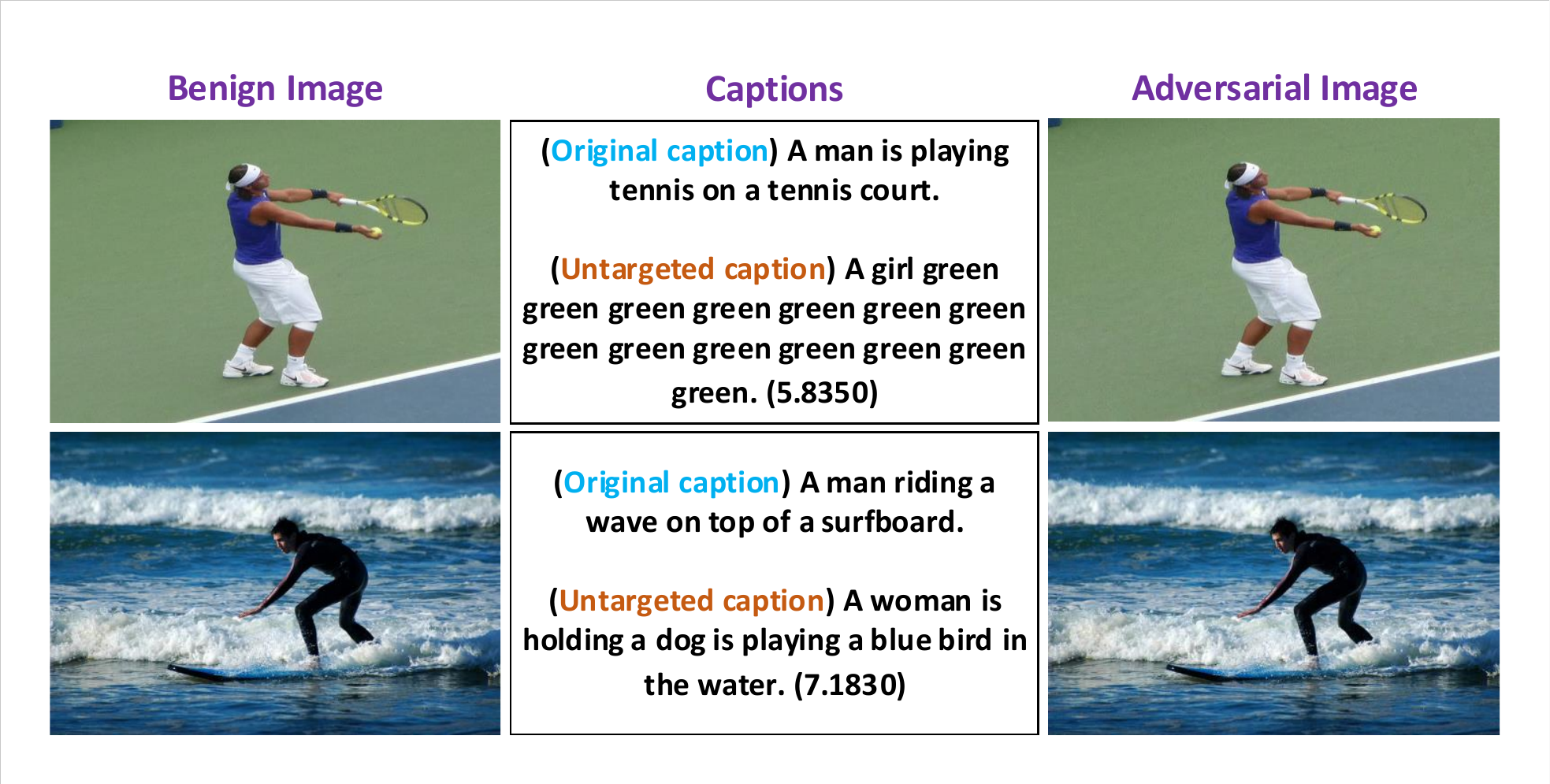}
\vspace{-1.6em}
\caption{Examples of untargeted attacks to Show-Attend-and-Tell.}
\label{fig: untargeted}
\vspace{-1.3em}
\end{figure}

\vspace{0.2em}
\noindent
{\bf Untargeted caption attack.} 
%In previous adversarial attacks to image classification, they often present both targeted and untargeted attacks. 
Until now, we have only presented targeted caption attacks. 
In the following, we present a brief analysis about the untargeted caption attack, which can be formulated as follows:
\vspace{-0.4em}
\begin{equation}
\arg\min_{\bepsilon} \ln P(\bS_0 | \bepsilon)  + \lambda \| \bepsilon \|_2^2, 
~  s.t. ~ \bI_0 + \bepsilon \in [0, 1],
\label{eq: min log likelihood of untargeted}
\vspace{-0.4em}
\end{equation}
where $\bS_0$ denotes the predicted caption on the benign image $\bI_0$. 
This problem can be easily solved by the projected gradient descent algorithm. 
Two attack results to the Show-Attend-and-Tell model are shown in Fig. \ref{fig: untargeted}. 
The predicted captions after attacking are non-meaningful, \ie, violating the grammar of natural language. 
It is not difficult to explain this observation. In image classification, the classification space is continuous and closed, and the prediction will jump from one to another label if the image is attacked. However, the distributions of meaningful captions are not continuous in image captioning. There are massive non-meaningful captions around every meaningful caption. 
Consequently, we think it makes no sense to calculate how many captions are fooled by untargeted attacks.
However, this simple analysis reveals an important information that state-of-the-art DNN-based image captioning systems have not learned or understood the grammar of natural language very well.  

\vspace{0.1em}
\noindent
{\bf A brief summary.} The above two studies demonstrate that state-of-the-art CNN+RNN image captioning systems are still far from human captioning.
The proposed methods can be used as a probe tool to check what grammars have been learned by the automatic image captioning system, thus to guide the improvement towards human captioning. 

%-------------------------------------------------------------------------
% \vspace{-0.3em}
\section{Conclusions} 
\label{sec: conclusion}
% \vspace{-0.25em}

In this paper, we have fooled the CNN+RNN based image captioning system to produce targeted partial captions by generating adversarial noises added onto benign images. 
We formulate the attack of targeted partial captions as a structured output learning problem. 
We further present two structured methods, including the generalized expectation maximization and the structural SVMs with latent variables. 
Extensive experiments demonstrate that state-of-the-art image captioning models 
%including Show-and-Tell, Show-Attend-and-Tell and SCST, 
can be easily attacked by the proposed methods. 
Furthermore, the proposed methods have been used to explore the inner mechanism of image caption systems, revealing that current automatic image captioning systems are far from human captioning.
In our future work, we plan to use the proposed methods to guide the improvement of automatic image captioning systems towards human captioning, and enhance the robustness. 
%\red{Moreover, the vulnerability revealed by attacking could provide the hint to enhance the robustness of automatic image captioning systems. It will be explored in our future work. }

\vspace{0.6em}
\noindent
{\bf Acknowledgement}. 
The involvements of Yan Xu, Fumin Shen and Heng Tao Shen in this work were supported in part by the National Natural Science Foundation of China under Project 61502081, Sichuan Science and Technology Program (No. 2019YFG0003, 2018GZDZX0032).

{\small
\bibliographystyle{ieee}
\bibliography{egbib}

\begin{thebibliography}{10}\itemsep=-1pt

\bibitem{attack-survey-2018}
N.~Akhtar and A.~Mian.
\newblock Threat of adversarial attacks on deep learning in computer vision: A
  survey.
\newblock {\em IEEE Access}, 6:14410--14430, 2018.

\bibitem{attack-segmentation-cvpr-2018}
A.~Arnab, O.~Miksik, and P.~H.~S. Torr.
\newblock On the robustness of semantic segmentation models to adversarial
  attacks.
\newblock In {\em Proceedings of the IEEE Conference on Computer Vision and
  Pattern Recognition}, 2018.

\bibitem{structured-output-learning-2007}
G.~BakIr, T.~Hofmann, B.~Sch{\"o}lkopf, A.~J. Smola, B.~Taskar, and
  S.~Vishwanathan.
\newblock {\em Predicting Structured Data}.
\newblock MIT press, 2007.

\bibitem{bishop-2006}
C.~Bishop.
\newblock {\em Pattern Recognition and Machine Learning}.
\newblock Springer, 2006.

\bibitem{CW-attack-2017}
N.~Carlini and D.~Wagner.
\newblock Towards evaluating the robustness of neural networks.
\newblock In {\em Proceedings of the IEEE Symposium on Security and Privacy},
  pages 39--57. IEEE, 2017.

\bibitem{attack-to-image-caption-acl-2018}
H.~Chen, H.~Zhang, P.-Y. Chen, J.~Yi, and C.-J. Hsieh.
\newblock Attacking visual language grounding with adversarial examples: A case
  study on neural image captioning.
\newblock In {\em Proceedings of the Association for Computational
  Linguistics}, volume~1, pages 2587--2597, 2018.

\bibitem{mi-fsgm-cvpr-2018}
Y.~Dong, F.~Liao, T.~Pang, H.~Su, J.~Zhu, X.~Hu, and J.~Li.
\newblock Boosting adversarial attacks with momentum.
\newblock In {\em Proceedings of the IEEE Conference on Computer Vision and
  Pattern Recognition}, 2016.

\bibitem{attack-to-detection-arxiv-2017}
K.~Eykholt, I.~Evtimov, E.~Fernandes, B.~Li, A.~Rahmati, F.~Tramer, A.~Prakash,
  T.~Kohno, and D.~Song.
\newblock Physical adversarial examples for object detectors.
\newblock {\em arXiv preprint arXiv:1807.07769}, 2018.

\bibitem{RCNN-CVPR-2014}
R.~Girshick, J.~Donahue, T.~Darrell, and J.~Malik.
\newblock Rich feature hierarchies for accurate object detection and semantic
  segmentation.
\newblock In {\em Proceedings of the IEEE Conference on Computer Vision and
  Pattern Recognition}, pages 580--587, 2014.

\bibitem{FSGM-ICLR-2015}
I.~Goodfellow, J.~Shlens, and C.~Szegedy.
\newblock Explaining and harnessing adversarial examples.
\newblock In {\em Proceedings of the International Conference on Learning
  Representations}, 2015.

\bibitem{resnet-kaiming-cvpr-2016}
K.~He, X.~Zhang, S.~Ren, and J.~Sun.
\newblock Deep residual learning for image recognition.
\newblock In {\em Proceedings of the IEEE Conference on Computer Vision and
  Pattern Recognition}, pages 770--778, 2016.

\bibitem{split-of-coco-2015}
A.~Karpathy and L.~Fei-Fei.
\newblock Deep visual-semantic alignments for generating image descriptions.
\newblock In {\em Proceedings of the IEEE Conference on Computer Vision and
  Pattern Recognition}, pages 3128--3137, 2015.

\bibitem{adam-2014}
D.~P. Kingma and J.~Ba.
\newblock Adam: A method for stochastic optimization.
\newblock {\em arXiv preprint arXiv:1412.6980}, 2014.

\bibitem{pgm-book-koller}
D.~Koller, N.~Friedman, and F.~Bach.
\newblock {\em Probabilistic Graphical Models: Principles and Techniques}.
\newblock MIT press, 2009.

\bibitem{alexnet-nips-2012}
A.~Krizhevsky, I.~Sutskever, and G.~E. Hinton.
\newblock Imagenet classification with deep convolutional neural networks.
\newblock In {\em Proceedings of the Advances in Neural Information Processing
  Systems}, pages 1097--1105, 2012.

\bibitem{I-FSGM-2016}
A.~Kurakin, I.~Goodfellow, and S.~Bengio.
\newblock Adversarial examples in the physical world.
\newblock {\em arXiv preprint arXiv:1607.02533}, 2016.

\bibitem{cnn-lecun-1995}
Y.~LeCun, Y.~Bengio, et~al.
\newblock Convolutional networks for images, speech, and time series.
\newblock {\em The Handbook of Brain Theory and Neural Networks},
  3361(10):1995, 1995.

\bibitem{deep-learning-nature-2015}
Y.~LeCun, Y.~Bengio, and G.~Hinton.
\newblock Deep learning.
\newblock {\em Nature}, 521(7553):436, 2015.

\bibitem{TADT}
X.~Li, C.~Ma, B.~Wu, Z.~He, and M.-H. Yang.
\newblock Target-aware deep tracking.
\newblock In {\em Proceedings of the IEEE Conference on Computer Vision and
  Pattern Recognition}, 2019.

\bibitem{mscoco-eccv-2014}
T.-Y. Lin, M.~Maire, S.~Belongie, J.~Hays, P.~Perona, D.~Ramanan,
  P.~Doll{\'a}r, and C.~L. Zitnick.
\newblock Microsoft coco: Common objects in context.
\newblock In {\em Proceedings of the European Conference on Computer Vision},
  pages 740--755. Springer, 2014.

\bibitem{fcn-segmentation-cvpr-2015}
J.~Long, E.~Shelhamer, and T.~Darrell.
\newblock Fully convolutional networks for semantic segmentation.
\newblock In {\em Proceedings of the IEEE Conference on Computer Vision and
  Pattern Recognition}, pages 3431--3440, 2015.

\bibitem{attack-to-object-detection-arxiv-2017}
J.~Lu, H.~Sibai, E.~Fabry, and D.~Forsyth.
\newblock No need to worry about adversarial examples in object detection in
  autonomous vehicles.
\newblock {\em arXiv preprint arXiv:1707.03501}, 2017.

\bibitem{deepfool-cvpr-2016}
S.-M. Moosavi-Dezfooli, A.~Fawzi, and P.~Frossard.
\newblock Deepfool: a simple and accurate method to fool deep neural networks.
\newblock In {\em Proceedings of the IEEE Conference on Computer Vision and
  Pattern Recognition}, pages 2574--2582, 2016.

\bibitem{self-critical-image-caption-cvpr-2017}
S.~J. Rennie, E.~Marcheret, Y.~Mroueh, J.~Ross, and V.~Goel.
\newblock Self-critical sequence training for image captioning.
\newblock In {\em Proceedings of the IEEE Conference on Computer Vision and
  Pattern Recognition}, 2017.

\bibitem{sgd-1985}
H.~Robbins and S.~Monro.
\newblock A stochastic approximation method.
\newblock In {\em Herbert Robbins Selected Papers}, pages 102--109. Springer,
  1985.

\bibitem{RNN-2015}
J.~Schmidhuber.
\newblock Deep learning in neural networks: An overview.
\newblock {\em Neural Networks}, 61:85--117, 2015.

\bibitem{VQA-iccv-2015}
A.~Stanislaw, A.~Aishwarya, L.~Jiasen, M.~Margaret, B.~Dhruv, Z.~C.~Lawrence,
  and P.~Devi.
\newblock Vqa: Visual question answering.
\newblock In {\em Proceedings of the IEEE International Conference on Computer
  Vision}, 2015.

\bibitem{inception-v3}
C.~Szegedy, V.~Vanhoucke, S.~Ioffe, J.~Shlens, and Z.~Wojna.
\newblock Rethinking the inception architecture for computer vision.
\newblock In {\em Proceedings of the IEEE Conference on Computer Vision and
  Pattern Recognition}, pages 2818--2826, 2016.

\bibitem{szegedy2013intriguing}
C.~Szegedy, W.~Zaremba, I.~Sutskever, J.~Bruna, D.~Erhan, I.~Goodfellow, and
  R.~Fergus.
\newblock Intriguing properties of neural networks.
\newblock {\em arXiv preprint arXiv:1312.6199}, 2013.

\bibitem{show-tell-cvpr-2015}
O.~Vinyals, A.~Toshev, S.~Bengio, and D.~Erhan.
\newblock Show and tell: A neural image caption generator.
\newblock In {\em Proceedings of the IEEE Conference on Computer Vision and
  Pattern Recognition}, pages 3156--3164, 2015.

\bibitem{tencent-ml-images-2018}
B.~Wu, W.~Chen, Y.~Fan, Y.~Zhang, J.~Hou, J.~Liu, J.~Huang, W.~Liu, and
  T.~Zhang.
\newblock Tencent ml-images: A large-scale multi-label image database for
  visual representation learning.
\newblock {\em arXiv preprint arXiv:1901.01703}, 2019.

\bibitem{my-cvpr-2018}
B.~Wu, W.~Chen, P.~Sun, W.~Liu, B.~Ghanem, and S.~Lyu.
\newblock Tagging like humans: Diverse and distinct image annotation.
\newblock In {\em Proceedings of the IEEE Conference on Computer Vision and
  Pattern Recognition}, pages 7967--7975, 2018.

\bibitem{my-cvpr-2017}
B.~Wu, F.~Jia, W.~Liu, and B.~Ghanem.
\newblock Diverse image annotation.
\newblock In {\em Proceedings of the IEEE Conference on Computer Vision and
  Pattern Recognition}, pages 2559--2567, 2017.

\bibitem{attack-to-segmentation-detection-iccv-2017}
C.~Xie, J.~Wang, Z.~Zhang, Y.~Zhou, L.~Xie, and A.~Yuille.
\newblock Adversarial examples for semantic segmentation and object detection.
\newblock In {\em Proceedings of the IEEE International Conference on Computer
  Vision}. IEEE, 2017.

\bibitem{show-attend-tell-icml-2015}
K.~Xu, J.~Ba, R.~Kiros, K.~Cho, A.~Courville, R.~Salakhudinov, R.~Zemel, and
  Y.~Bengio.
\newblock Show, attend and tell: Neural image caption generation with visual
  attention.
\newblock In {\em Proceedings of the International Conference on Machine
  Learning}, pages 2048--2057, 2015.

\bibitem{xu_2018_CVPR}
X.~Xu, X.~Chen, C.~Liu, A.~Rohrbach, T.~Darrell, and D.~Song.
\newblock Fooling vision and language models despite localization and attention
  mechanism.
\newblock In {\em Proceedings of the IEEE Conference on Computer Vision and
  Pattern Recognition}, June 2018.

\bibitem{ssvm-latent-icml-2009}
C.-N.~J. Yu and T.~Joachims.
\newblock Learning structural svms with latent variables.
\newblock In {\em Proceedings of the International Conference on Machine
  Learning}, pages 1169--1176. ACM, 2009.

\bibitem{zhang2012robust}
T.~Zhang, B.~Ghanem, S.~Liu, and N.~Ahuja.
\newblock Robust visual tracking via multi-task sparse learning.
\newblock In {\em Proceedings of the IEEE Conference on Computer Vision and
  Pattern Recognition}, pages 2042--2049, 2012.

\end{thebibliography}
}

\end{document}